\documentclass[runningheads]{llncs}

\usepackage{eccv}

\usepackage{eccvabbrv}
\usepackage{comment}
\usepackage{siunitx}
\usepackage{xcolor}
\usepackage{tcolorbox}

\usepackage{graphicx}
\usepackage{wrapfig}
\usepackage{booktabs}
\usepackage{multirow}
\usepackage[accsupp]{axessibility}  

\usepackage{orcidlink}

\begin{document}

\title{Reference-Free Image Quality Assessment for Virtual Try-On via Human Feedback} 

\titlerunning{Reference-Free Image Quality Assessment for VTON via Human Feedback}

\author{Yuki Hirakawa\inst{1,3} \and Takashi Wada\inst{1} \and Ryotaro Shimizu\inst{1} \and Takuya Furusawa\inst{1} \and Yuki Saito\inst{1} \and Ryosuke Araki\inst{2} \and Tianwei Chen\inst{1} \and Fan Mo\inst{1} \and Yoshimitsu Aoki\inst{3}}

\authorrunning{Y.~Hirakawa et al.}

\institute{
ZOZO Research \and
ZOZO Inc.  \and
Keio University
}

\maketitle

\begin{abstract}
As virtual try-on (VTON) systems become increasingly important in fashion e-commerce, there is a growing need for reliable reference-free evaluation methods, since ground-truth images of the same person wearing the target garment are typically unavailable in real-world scenarios. To address this challenge, we propose VTON-IQA, a reference-free framework for human-aligned image quality assessment without requiring ground-truth images. To model human perceptual judgments, we construct VTON-QBench, a large-scale human-annotated benchmark comprising 62,688 try-on images generated by 14 representative VTON models and 431,800 quality annotations collected from 13,838 qualified annotators. To the best of our knowledge, this is the largest dataset to date for human subjective evaluation in VTON. Extensive experiments show that VTON-IQA achieves reliable human-aligned image quality assessment. Moreover, we conduct a comprehensive benchmark evaluation of 14 representative VTON models using VTON-IQA. The dataset and source code are released at \url{https://github.com/litelightlite/VTON-IQA}.
\end{abstract}
\section{Introduction}
Given a person image and a garment image, image-based virtual try-on (VTON) synthesizes a try-on image of the person wearing the target garment. In the fashion e-commerce domain, it has attracted considerable attention as a means of providing an online try-on experience and bridging the gap between online and in-store shopping. Despite its growing practical importance, reliable evaluation remains a fundamental challenge. In real-world deployment, ground-truth images of the same person wearing the target garment are typically unavailable, rendering reference-based metrics such as Structural Similarity Index (SSIM)~\cite{ssim} and Learned Perceptual Image Patch Similarity (LPIPS)~\cite{lpips} impractical. Although large-scale user studies can offer reliable assessments, they are costly, time-consuming, and difficult to deploy in production settings. Distribution-level metrics such as Fréchet Inception Distance (FID)~\cite{fid} and Kernel Inception Distance (KID)~\cite{kid}, on the other hand, capture dataset-level statistics but fail to assess the perceptual quality of individual generated images. Recent efforts have introduced virtual try-on evaluation methods that do not rely on ground-truth images~\cite{vtonqa, vtonvllm, vtbench}. While these methods mark important progress toward task-specific, reference-free evaluation, their alignment with large-scale human judgments remains insufficiently validated. Moreover, the lack of publicly available implementations and standardized benchmarks hinders reproducible evaluation and limits sustained progress in the virtual try-on community.

To address these challenges, we propose VTON-IQA, a reference-free evaluation framework that performs image quality assessment (IQA) of virtual try-on images without relying on ground truth. Given a person image, a garment image, and the corresponding generated try-on image, VTON-IQA predicts a continuous quality score aligned with human subjective evaluations. To model human perceptual judgments, we construct VTON-QBench, a large-scale human-annotated benchmark tailored to virtual try-on images. VTON-QBench consists of 62,688 try-on images generated by 14 representative VTON models and annotated through crowdsourced subjective evaluations. Since crowdsourcing inevitably introduces variability in annotation quality, including inattentive or adversarial responses, we design a dedicated data curation pipeline to quantitatively assess annotator consistency and reliability. This process allows us to identify and exclude unreliable annotators from the final dataset. As a result, VTON-QBench contains 431,800 high-quality annotations collected from 13,838 qualified annotators. To the best of our knowledge, this is the largest dataset to date for human subjective quality evaluation in the virtual try-on domain. With VTON-IQA and VTON-QBench, we provide a scalable and reproducible alternative to large-scale user studies for evaluating virtual try-on quality.

In virtual try-on, quality assessment must account for not only the perceptual quality of the generated image, but also garment fidelity and the preservation of non-target visual elements, such as the person’s identity, body shape, pose, and background. This requirement fundamentally distinguishes try-on image assessment from conventional single-image quality assessment, which predicts quality solely from a single input image~\cite{qalign, clipiqa}. Accordingly, evaluating try-on images requires modeling cross-image interactions among the generated try-on image, the input garment image, and the source person image. To this end, we introduce an Interleaved Cross-Attention (ICA) module, which extends standard transformer blocks by inserting a cross-attention layer between the self-attention and MLP layers in the latter half of the blocks. By enabling structured interactions across the try-on, garment, and person representations, ICA allows the model to jointly assess whether garment attributes are faithfully transferred and whether non-target visual elements remain visually consistent. Extensive experiments demonstrate that ICA effectively captures quality factors specific to virtual try-on and achieves strong alignment with human subjective judgments.

Furthermore, we conduct a comprehensive benchmark evaluation of 14 representative virtual try-on models using VTON-IQA. Beyond these baseline evaluations, the public release of VTON-IQA can provide a fair, reproducible, and reference-free evaluation criterion for future virtual try-on methods, serving as a reproducible alternative to the ad hoc user studies often conducted in individual research projects. We expect that this will contribute to the standardization of quality assessment and further advance research in virtual try-on.
\section{Related Work}
\noindent\textbf{Image-based Virtual Try-On.}
Given a person image and a garment image, image-based virtual try-on synthesizes a try-on image of the person wearing the target garment. Early approaches~\cite{viton, vitonhd, hrviton, sdviton} predominantly followed a two-stage pipeline. A clothing-agnostic representation was first constructed by removing garment-related information from the person image, after which the target garment was geometrically aligned to the body using Thin-Plate Spline transformation and fused via a Generative Adversarial Network (GAN)~\cite{gan}. Subsequent works improved robustness to complex poses and occlusions through human parsing, pose estimation, and enhanced warping–fusion modules, leading to better texture preservation and more natural garment boundaries~\cite{vitonhd, hrviton, sdviton, clothflow, acgpn, pfafn, gpvton}. More recently, diffusion-based approaches have gained popularity due to their strong ability to reproduce high-frequency details and generate high-resolution images~\cite{ddpm, stablediffusion}. Early diffusion-based VTON models leveraged pretrained latent diffusion models and incorporated garment features as conditioning signals within U-Net architectures~\cite{tryondiffusion, ladi}. Later works explored conditioning strategies and architectural refinements. For example, IDM-VTON~\cite{idm} enhances garment fidelity by disentangling semantic and texture features within attention modules, whereas CatVTON~\cite{catvton} adopts a simplified end-to-end design by spatially concatenating person and garment images. Recent efforts have further transitioned to Diffusion Transformers (DiT)~\cite{dit}, which offer stronger global modeling through self-attention~\cite{any2any, fitdit}, enabling improved long-range dependency modeling of garment structure and texture. In parallel, advances in proprietary image editing models~\cite{nanobanana, gptimage} demonstrate that high-quality virtual try-on can be achieved in a zero-shot manner using a person image, a garment image, and appropriate textual prompts.

\noindent\textbf{Quality Assessment for Virtual Try-On.}
Recent works have explored evaluation frameworks for virtual try-on that do not require ground-truth images. VTON-VLLM~\cite{vtonvllm} trains a multimodal large language model using human annotations on LLM-generated critiques of synthesized try-on images, enabling preference-aware evaluation and feedback. However, its focus is on validating and generating textual critiques rather than learning a direct quantitative image-level quality predictor. VTBench~\cite{vtbench} introduces a hierarchical benchmark that evaluates virtual try-on results across multiple quality dimensions. Although it incorporates human-aligned signals through LLM-based general aesthetic predictor, it does not use a quality assessment model specifically tailored to virtual try-on. VTONQA~\cite{vtonqa} is most closely related to our work, as it constructs a human-annotated dataset to train a virtual try-on quality evaluator. However, its relatively limited dataset scale may restrict the robustness of the learned model. In contrast, we construct a large-scale human-annotated benchmark and learn a reference-free, image quality assessment model aligned with human perceptual judgments.
\section{VTON-QBench}
\begin{figure}[t]
    \centering
    \includegraphics[width=\linewidth]{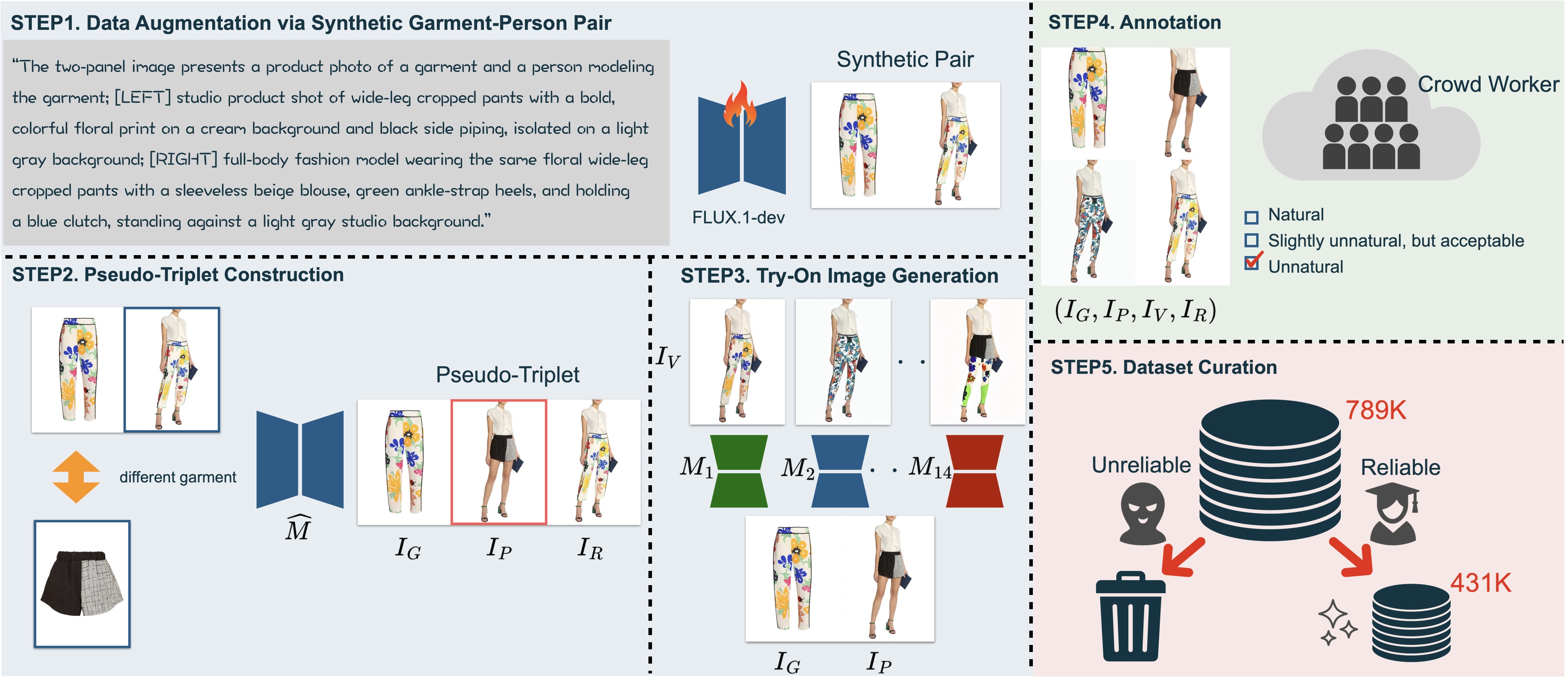}
    \caption{Overview of the VTON-QBench construction pipeline. VTON-QBench is built through five stages: (1) synthetic garment–person pair augmentation via FLUX.1-dev, (2) pseudo-triplet construction, (3) virtual try-on image generation using 14 representative VTON models, (4) crowdsourced human annotation with reference images, and (5) dataset curation to remove unreliable annotations. This pipeline ensures fashion diversity, controlled evaluation settings, and reliable human-aligned quality labels.}
    \label{fig:overview}
\end{figure}
\begin{table}[t]
\centering
\caption{Comparison of dataset statistics. VTON-QBench significantly expands the scale of garment–person pairs, try-on images, annotators, and quality annotations compared to prior work, and is publicly available (OSS).}
\label{tab:dataset_comparison}
\begin{tabular}{lrrrrrr}
\toprule
Method 
& \# Pair 
& \# Try-On 
& \# VTON Model 
& \# Annotator
& \# Annotation
& OSS \\
\midrule
VTONQA~\cite{vtonqa} 
& 748 & 8,132 & 11 & 40 & N/R & No \\\midrule
VTON-QBench
& 13,153 & 62,688 & 14 & 13,838 & 431,800 & Yes \\
\bottomrule
\end{tabular}
\end{table}
To enable human-aligned quality assessment for VTON, we construct VTON-QBench, a large-scale dataset comprising 62,688 try-on images generated by 14 representative models and 431,800 quality annotations collected from 13,838 crowd workers who met reliability criteria. Fig.~\ref{fig:overview} shows an overview of the VTON-QBench construction pipeline, which consists of five stages: (1) synthetic garment–person pair augmentation via FLUX.1-dev, (2) pseudo-triplet construction, (3) virtual try-on image generation using 14 representative VTON models, (4) crowdsourced human annotation with reference images, and (5) dataset curation to remove unreliable annotations. As shown in Table~\ref{tab:dataset_comparison}, VTON-QBench is larger in scale than the concurrent work~\cite{vtonqa} and, to the best of our knowledge, the only publicly available dataset for human subjective evaluation in VTON.
\subsection{Image Preparation}
\begin{figure}[t]
  \centering
  \begin{minipage}[t]{0.49\linewidth}
    \centering
    \includegraphics[height=4.0cm]{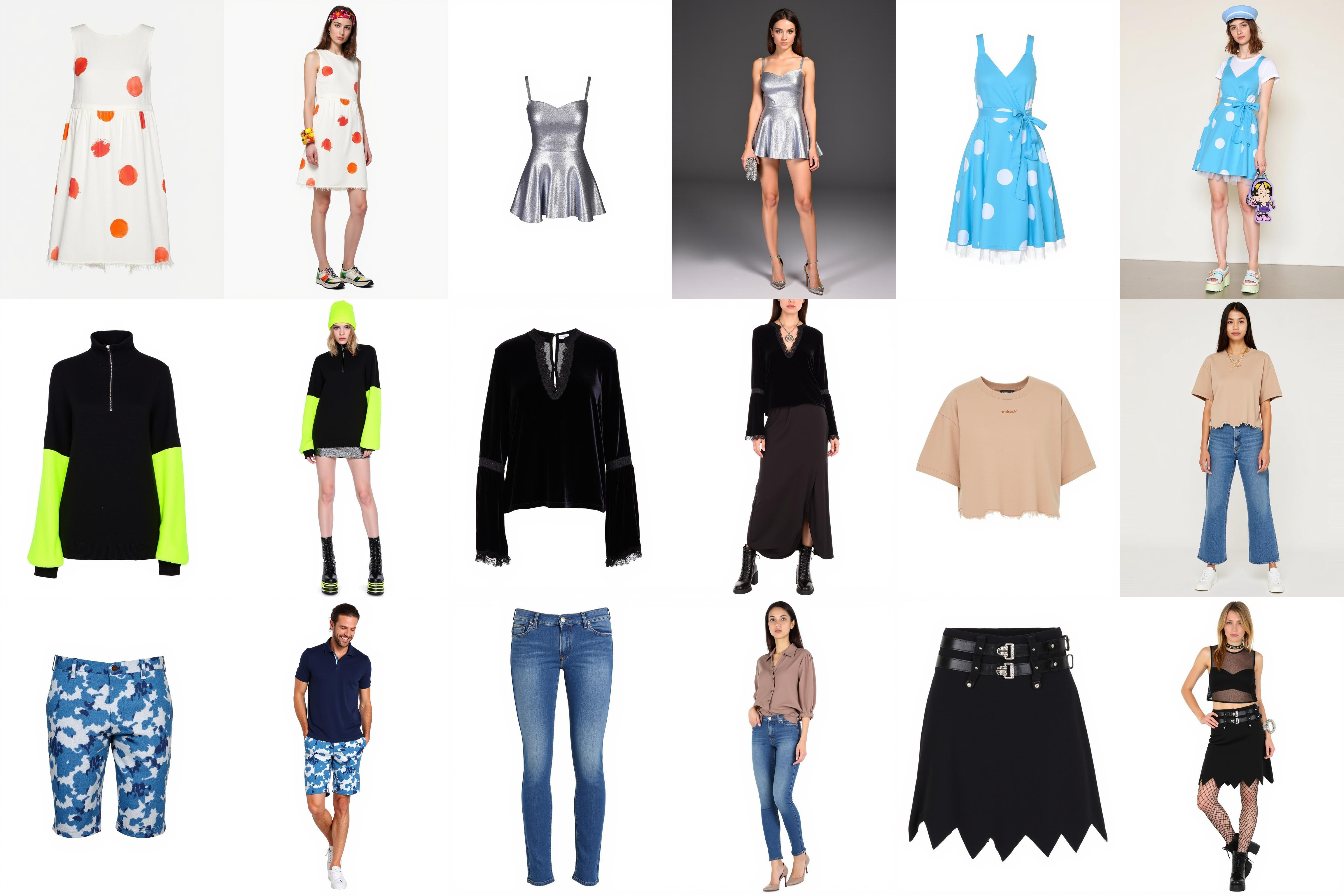}
    \subcaption{Synthetic garment–person pairs.}
    \label{fig:synth_vton}
  \end{minipage}
  \hfill
  \begin{minipage}[t]{0.49\linewidth}
    \centering
    \includegraphics[height=4.0cm]{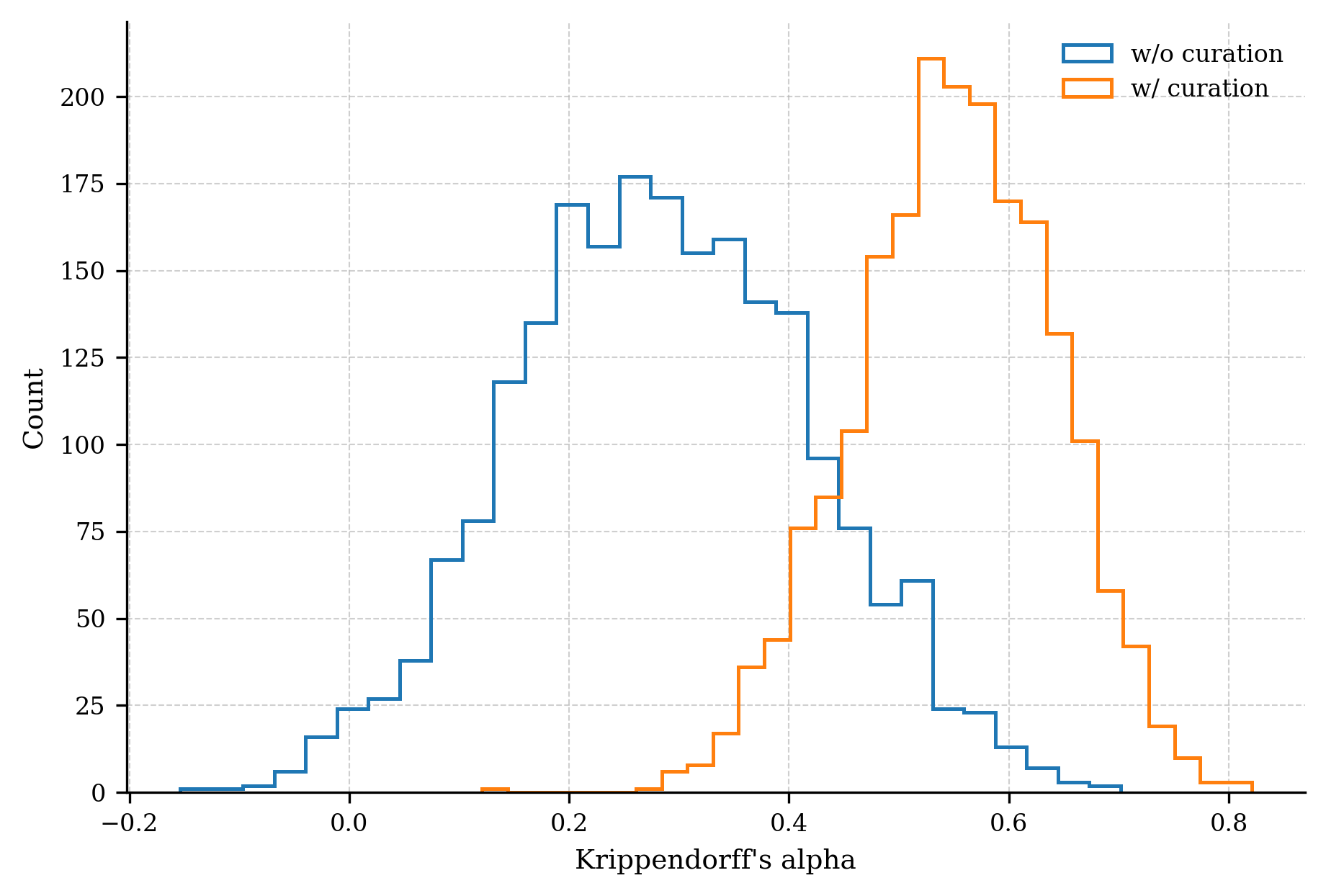}
    \subcaption{Distribution of Krippendorff’s $\alpha$.}
    \label{fig:krippendorff}
  \end{minipage}
  \caption{Left: Representative synthetic garment–person pairs introduced to enhance fashion diversity. Right: Distribution of Krippendorff’s $\alpha$ before and after data curation, showing improved inter-annotator agreement after filtering unreliable annotations.}
  \label{fig:dataset_curation}
\end{figure}

\noindent{\textbf{Data Augmentation via Synthetic Garment--Person Pairs.}} 
VTON-QBench is constructed from the test splits of VITON-HD~\cite{vitonhd} and Dress Code~\cite{dresscode}, both of which provide high-quality garment--person image pairs $(I_G, I_P)$. However, due to the limited diversity of their test sets, the constructed pairs lack variation in garment styles, body shapes, poses, and person appearances. To improve coverage beyond the original benchmarks, we augment the data with synthetic garment--person pairs, as illustrated in Fig.~\ref{fig:synth_vton}. Specifically, we generate additional pairs according to the fashion style taxonomy of Mystylebox~\cite{mystylebox2026fashion}, covering 24 diverse styles. For this generation process, we employ FLUX.1-dev~\cite{flux2024} and train a LoRA specialized for garment--person pair synthesis~\cite{lhhuang2024iclora} using an internally collected dataset. This allows us to obtain paired garment and person images with broader variations in appearance while maintaining compatibility between the garment and the target person. Since synthetic generation may introduce inconsistencies between the garment image and the corresponding person image, such as mismatches in category, color, silhouette, or texture, we apply a two-stage verification process. First, we perform automatic attribute-consistency filtering using GPT-4.1~\cite{gpt4} to remove pairs with obvious semantic or visual mismatches. Second, experienced reviewers familiar with fashion images, including researchers in related computer vision fields, manually inspect the remaining pairs to ensure that garment attributes are preserved and that the generated person images are visually plausible. We retain only pairs that pass both stages. As a result, the number of garment--person pairs increases from 6,981 to 13,153, corresponding to approximately 1.9$\times$ the original scale. Additional details on the generation procedure are provided in the supplementary material.

\noindent{\textbf{Pseudo-Triplet Construction.}} 
To enable comparison with existing reference-based metrics such as SSIM~\cite{ssim} and LPIPS~\cite{lpips}, we construct pseudo-triplets from existing garment--person pairs $(I_{G'}, I_{P'})$, where $I_{P'}$ is the ground-truth person image wearing garment $I_{G'}$. Given a pair $(I_{G'}, I_{P'})$, we sample a different garment image $I_{G''}$ such that $G'' \neq G'$, and apply a strong virtual try-on model $\widehat{M}$ to generate an intermediate try-on image $I_{V''} = \widehat{M}(I_{G''}, I_{P'})$, in which the original garment $G'$ is replaced with $G''$. We then define the pseudo-triplet as $(I_G, I_P, I_R) = (I_{G'}, I_{V''}, I_{P'})$, where $I_{G'}$ serves as the target garment image, $I_{V''}$ as the input person image, and $I_{P'}$ as the ground-truth reference image. Given this pseudo-triplet, a target VTON model $M$ generates try-on image $I_V = M(I_{G'}, I_{V''})$, and reference-based metrics can be computed by comparing $I_V$ and $I_{P'}$. This construction converts an otherwise unpaired setting into a pseudo-reference setting, allowing systematic comparison with reference-based metrics. In our experiments, we use Nano Banana Pro~\cite{nanobanana} as the strong virtual try-on model $\widehat{M}$. Since generating $I_{V''}$ may introduce unintended changes, such as pose shifts, background alterations, or modifications outside the target garment region, we manually filter the generated samples and retain only those in which garment replacement is the primary difference while the person structure and background are preserved.

\noindent{\textbf{Try-On Image Generation.}} 
We generate virtual try-on images $I_V$ using 14 representative models: VITON-HD~\cite{vitonhd}, HR-VITON~\cite{hrviton}, LADI-VTON~\cite{ladi}, SD-VITON~\cite{sdviton}, CAT-DM~\cite{catdm}, OOTDiffusion~\cite{oot}, IDM-VTON~\cite{idm}, CatVTON~\cite{catvton}, CatVTON-FLUX~\cite{catvton}, FitDit~\cite{fitdit}, Any2AnyTryon~\cite{any2any}, Qwen-Image-Edit~\cite{qwenedit}, Nano Banana Pro~\cite{nanobanana}, and GPT-Image-1.5~\cite{gptimage}. For each model, we use publicly released pretrained weights and follow the recommended inference settings whenever available. 
\subsection{Annotation Protocol}
For each clothing image $I_G$, person image $I_P$, and generated try-on image $I_V$, we collect independent quality annotations $\mathcal{A}$ from multiple crowd workers. Our preliminary experiments showed that garment length consistency is difficult to assess from the clothing image $I_G$ alone. Therefore, we additionally provide a reference image $I_R$, where the target garment is worn by a real person, as auxiliary context. Each questionnaire contains 50 evaluation tasks with a fixed structure and rating format. By varying the presented image tuples, we construct 2,139 questionnaires, each of which is assigned to multiple crowd workers to obtain repeated independent evaluations for every sample. 
Each annotation $a \in \mathcal{A}$ is given on a three-level ordinal scale: 
(1) unnatural, 
(2) slightly unnatural but acceptable, and 
(3) natural. 
The final subjective quality score $S(I_G, I_P, I_V)$ is computed as the average numerical rating across annotators:
\begin{equation}
S(I_G,I_P,I_V)
=
\frac{1}{\left| \mathcal{A}(I_G,I_P,I_V) \right|}
\sum_{a \in \mathcal{A}(I_G,I_P,I_V)} L(a),
\end{equation}
\noindent
where $L(a)$ is defined as follows:
\begin{equation}
L(a)=
\begin{cases}
1 & \text{if } a=\text{``Unnatural''},\\
2 & \text{if } a=\text{``Slightly unnatural, but acceptable''},\\
3 & \text{if } a=\text{``Natural''}.\\
\end{cases}
\end{equation}
\subsection{Dataset Curation}
Crowdsourced annotation inevitably introduces variability in response quality, including careless or inconsistent responses. To collect reliable annotations, we design a two-stage curation pipeline that removes unreliable annotators and low-agreement annotations. First, we perform annotator-level filtering using five shared dummy tasks with unambiguous answers, which are included among the 50 tasks assigned to each annotator. Annotators who fail these sanity checks are removed. We further exclude annotators who select identical responses for more than 80\% of the tasks or disagree with the majority vote in more than 60\% of cases. These annotator-level filters reduce the total number of annotations from 1,149,144 to 480,412. Next, we perform questionnaire-level filtering based on inter-rater agreement. Specifically, we compute Krippendorff's $\alpha$~\cite{krippendorff} for the three-level ordinal labels within each questionnaire and discard questionnaires with $\alpha \leq 0.4$. This yields 431,800 retained annotations after curation. As shown in Fig.~\ref{fig:krippendorff}, the average agreement increases from 0.286 to 0.550, and 94.5\% of the retained questionnaires achieve $\alpha > 0.4$. Given the inherent variability of subjective evaluation, an $\alpha$ above 0.4 is generally considered acceptable, indicating substantially improved annotation consistency~\cite{ku2024imagenhub}.
\section{Image Quality Assessment for VTON (VTON-IQA)}
Given a garment image $I_G$, a person image $I_P$, and a generated try-on image $I_V$, VTON-IQA predicts a continuous quality score $\hat{s}\in[-1, 1]$ aligned with human perceptual judgments. To effectively assess try-on quality from these inputs, we introduce the Interleaved Cross-Attention (ICA) module, which captures cross-image relationships among the garment, person, and try-on images. This section presents the overall architecture of VTON-IQA, followed by a detailed description of the proposed ICA module.
\begin{figure}[t]
    \centering
    \includegraphics[width=\linewidth]{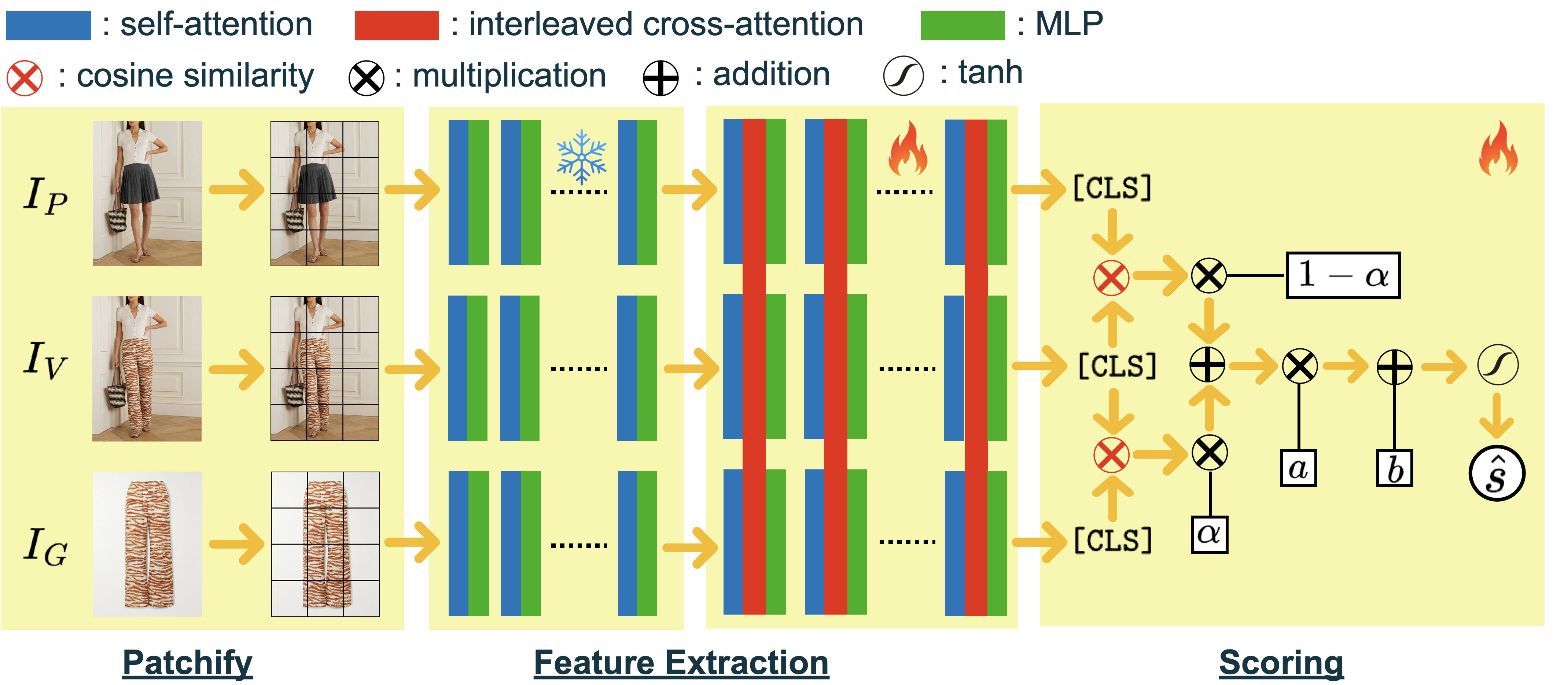}
    \caption{Architecture of VTON-IQA. The network processes $I_G$, $I_P$, and $I_V$ through a three-branch transformer backbone. The first half of layers perform independent feature extraction, while the latter half incorporates Interleaved Cross-Attention (ICA) to explicitly model cross-image interactions. The scoring module aggregates \texttt{[CLS]} representations to predict a human-aligned image-level quality score.}
    \label{fig:architecture}
\end{figure}
\subsection{Model Architecture}
We adopt a three-branch transformer architecture to process the garment image $I_G$, the person image $I_P$, and the generated try-on image $I_V$, as illustrated in Fig.~\ref{fig:architecture}. Each image is first embedded into a sequence of patch tokens augmented with a \texttt{[CLS]} token:
\begin{equation}
X_m^{(0)} = \phi(I_m) \in \mathbb{R}^{N \times d},
\quad m \in \{G,P,V\},
\end{equation}
where $N$ is the number of tokens and $d$ is the embedding dimension. The feature extraction module consists of $L$ transformer blocks. For the first half of the network, standard transformer blocks are applied independently to each branch:
\begin{equation}
\begin{aligned}
\widetilde{X}_m^{(\ell)} 
&= X_m^{(\ell-1)} + \mathrm{SA}^{(\ell)}(X_m^{(\ell-1)}), \\
X_m^{(\ell)} 
&= \widetilde{X}_m^{(\ell)} + \mathrm{MLP}^{(\ell)}(\widetilde{X}_m^{(\ell)}),
\end{aligned}
\quad \ell=1,\dots,L/2,\quad m \in \{G,P,V\}.
\end{equation}
In the latter $L/2$ layers, we extend the standard transformer block by introducing the proposed ICA module. Unlike conventional multi-branch designs that symmetrically model all pairwise interactions, ICA is motivated by the observation that virtual try-on quality primarily depends on the consistency between the generated try-on image and its corresponding garment and person images. Specifically, after the self-attention operation, cross-attention is applied between the try-on representation and each of the garment and person representations. In other words, interactions are explicitly modeled for the modality pairs $(V,G)$ and $(V,P)$ in both directions. The cross-attention features are computed as
\begin{equation}
\begin{aligned}
C_{m \leftarrow n}^{(\ell)} 
&= \mathrm{CA}^{(\ell)}(
Q=\widetilde{X}_m^{(\ell)},
K=\widetilde{X}_n^{(\ell)},
V=\widetilde{X}_n^{(\ell)}
), \\
&\quad (m,n) \in \{(V,G),(G,V),(V,P),(P,V)\}.
\end{aligned}
\end{equation}
To reflect the asymmetric dependency structure, the try-on representation $X_V$ aggregates contributions from both garment and person features:
\begin{equation}
\widehat{X}_V^{(\ell)} =
\widetilde{X}_V^{(\ell)}
+
C_{V\leftarrow G}^{(\ell)}
+
C_{V\leftarrow P}^{(\ell)}.
\end{equation}
In contrast, the garment and person branches only incorporate information from the try-on branch:
\begin{equation}
\widehat{X}_G^{(\ell)} = \widetilde{X}_G^{(\ell)} + C_{G \leftarrow V}^{(\ell)}, \quad
\widehat{X}_P^{(\ell)} = \widetilde{X}_P^{(\ell)} + C_{P \leftarrow V}^{(\ell)}.
\end{equation}
This asymmetric interaction design emphasizes that the quality judgment is fundamentally centered on the generated try-on image, which must be evaluated with respect to both garment fidelity and preservation of non-target visual elements. By explicitly modeling $V \leftrightarrow G$ and $V \leftrightarrow P$ interactions while avoiding unnecessary $G \leftrightarrow P$ coupling, ICA provides structured relational modeling tailored to virtual try-on quality assessment. After the final layer $L$, we extract the \texttt{[CLS]} token from each branch to obtain compact global representations $c_G$, $c_P$, and $c_V$ for the garment, person, and try-on images, respectively. We first compute an intermediate relational score $\tilde{s}$ as a weighted combination of cosine similarities between the try-on representation and the garment/person representations:
\begin{equation}
\tilde{s}
=
\alpha \frac{c_G^\top c_V}{\|c_G\| \, \|c_V\|}
+
(1-\alpha)\frac{c_P^\top c_V}{\|c_P\| \, \|c_V\|},
\label{cossim}
\end{equation}
where $\alpha \in [0,1]$ is a learnable scalar parameter that adaptively balances the relative importance of garment consistency and preservation of non-target regions. This formulation explicitly captures two complementary aspects of virtual try-on quality: (i) garment consistency via the similarity between $c_G$ and $c_V$, and (ii) preservation of non-target visual elements via the similarity between $c_P$ and $c_V$. By learning $\alpha$, the model dynamically weights these two relational components, allowing the overall quality score to reflect their relative significance in human perceptual judgment. Finally, the predicted score is obtained through a learnable affine transformation followed by a $\tanh$ activation:
\begin{equation}
\hat{s} = \tanh(a \tilde{s} + b),
\label{tanh}
\end{equation}
where $a$ and $b$ are learnable scalar parameters. The $\tanh$ function constrains the score to the bounded interval $[-1,1]$, which improves numerical stability, enhances interpretability, and enables consistent comparison across models.
\subsection{Loss Function}
Human ratings for virtual try-on images can exhibit high variance on an absolute scale, whereas relative preferences between two outputs for the same person–garment pair are often more consistent. We therefore optimize a joint objective that combines pairwise preference learning with score regression. Inspired by the Bradley–Terry model~\cite{bt}, we model pairwise preferences between two try-on results, $I_{V_i}$ and $I_{V_j}$, generated from the same person–garment pair $(I_G, I_P)$. The predicted preference probability is defined as
\begin{equation}
p_{\theta}
:=
P_{\theta}(I_{V_i} \succ I_{V_j} \mid I_G, I_P)
=
\sigma\!\left(
\frac{
\Psi_{\theta}(I_G, I_P, I_{V_i})
-
\Psi_{\theta}(I_G, I_P, I_{V_j})
}{\tau}
\right),
\end{equation}
where $\sigma(\cdot)$ denotes the sigmoid function, $\Psi_{\theta}(\cdot)$ is the predicted quality score, and $\tau$ is a temperature parameter. Similarly, given the human evaluation scores $S_i$ and $S_j$ assigned to $I_{V_i}$ and $I_{V_j}$, the empirical human preference probability is defined as
\begin{equation}
q_{ij}
:=
P_{\mathrm{human}}(I_{V_i} \succ I_{V_j} \mid I_G, I_P)
=
\sigma\!\left(
\frac{S_i - S_j}{\tau}
\right).
\end{equation}
The overall objective is given by the soft-label cross-entropy
combined with a score regression term:
\begin{equation}
\mathcal{L}_{\theta}
=
-\, q_{ij}\log p_{\theta}
-
(1-q_{ij})\log(1-p_{\theta})
+
\sum_{k\in\{i,j\}}
\left\|
\Psi_{\theta}(I_G, I_P, I_{V_k}) - S_k
\right\|_2^2.
\end{equation}
The first term aligns the predicted pairwise preference distribution
with the empirical human preference distribution via soft-label
cross-entropy, while the second term enforces consistency between
the predicted quality scores and the corresponding human ratings.
\section{Experiments}
We train VTON-IQA on VTON-QBench and evaluate it on the held-out test set against conventional reference-based metrics and a zero-shot baseline model. We report correlation with human ratings and pairwise ranking accuracy, and conduct ablation studies to quantify the effect of the proposed ICA module. Finally, we apply VTON-IQA to benchmark 14 representative VTON models on Dress Code and VITON-HD under both paired and unpaired settings.

\subsection{Experimental Setup}
\noindent{\textbf{Implementation Details.}} We build VTON-IQA on top of DINOv3 ViT-L/16~\cite{dinov3}. To integrate the proposed ICA module, we augment the last 12 transformer blocks with ICA layers and fine-tune both the original parameters of these blocks and the inserted ICA layers, while freezing the parameters of the first 12 layers. The model is optimized using AdamW~\cite{adamw} with a batch size of 16 and a learning rate of $1\times10^{-4}$. We employ early stopping based on the validation loss, selecting the checkpoint with the lowest validation loss if no improvement is observed for three consecutive epochs. Training is conducted on a single NVIDIA A100 (40GB) GPU using bfloat16 mixed-precision. 

\noindent{\textbf{Dataset.}} We split VTON-QBench into 43,948 training, 5,702 validation, and 13,038 test samples, with disjoint person and garment identities across splits. The quality scores are normalized to $[-1.0, 1.0]$ using Min--Max normalization followed by a linear transformation, with the normalization parameters estimated from the training set and applied consistently to the validation and test sets.

\noindent\textbf{Baselines.}
We compare our method with representative reference-based metrics, including SSIM~\cite{ssim} and LPIPS~\cite{lpips}. As a strong reference-free baseline, we include zero-shot DINOv3~\cite{dinov3}, in which garment, person, and try-on images are processed independently and the final score is computed using the cosine similarity formulation in Eq.~\eqref{cossim}, with $\alpha=0.5$. To assess the effectiveness of ICA, we further evaluate VTON-IQA with and without ICA.

\noindent{\textbf{Metrics.}} We evaluate prediction performance by measuring agreement with human subjective scores. We report Pearson's Linear Correlation Coefficient (PLCC, $\rho_{\mathrm{PLCC}}$) and Spearman's Rank Correlation Coefficient (SRCC, $\rho_{\mathrm{SRCC}}$) over all generated try-on images to assess global score consistency with human ratings. PLCC measures linear agreement, while SRCC evaluates whether the predicted scores preserve the ranking induced by human judgments. We also report the coefficient of determination ($R^2$) to assess regression accuracy with respect to absolute human scores. We further evaluate within-pair ranking consistency using pairwise accuracy at the garment--person pair level. For each garment--person pair $M$, all generated try-on images are compared pairwise, resulting in $\binom{N_M}{2}$ comparisons, where $N_M$ is the number of images for pair $M$. A comparison is counted as correct if the predicted scores preserve the relative ordering of the corresponding human scores. We report micro-averaged pairwise accuracy ($A_{\rm micro}$), computed over all pairwise comparisons, and macro-averaged pairwise accuracy ($A_{\rm macro}$), computed per pair and then averaged. Because $N_M$ is relatively small and varies across pairs, $A_{\rm macro}$ can be unstable; thus, we use $A_{\rm micro}$ as the main pairwise metric and $A_{\rm macro}$ as a complementary metric.
\begin{table}[t]
\centering
\caption{Comparison with baseline performance. Note that LPIPS is originally a ``lower-is-better'' metric; for consistency, we compute all metrics on its negated values. For SSIM, LPIPS, and zero-shot DINOv3, $\rho_{\mathrm{PLCC}}$ and $R^{2}$ are not reported, as their score scales are not directly comparable to human subjective ratings.}
\label{tab:compare_with_baseline}
\begin{tabular}{lccccc|cc}
\toprule
Scorer 
& {$\rho_{\mathrm{SRCC}} \uparrow$}
& {$\rho_{\mathrm{PLCC}} \uparrow$}
& {$R^2 \uparrow$}
& {$A_{\rm macro} \uparrow$}
& {$A_{\rm micro} \uparrow$}
& reference-free 
& fine-tuned\\
\midrule
SSIM~\cite{ssim}
& 0.150 & -- & -- & 0.596 & 0.593 & $\times$ & $\times$\\
LPIPS~\cite{lpips}
& 0.406 & -- & -- & 0.701 & 0.695 & $\times$ & $\times$\\
DINOv3~\cite{dinov3}
& 0.244 & -- & -- & 0.637 & 0.641 & $\checkmark$ & $\times$\\
\midrule
VTON-IQA w/o ICA
& 0.617 & 0.615 & 0.372 & 0.722 & 0.747 & $\checkmark$ & $\checkmark$\\
VTON-IQA
& \textbf{0.750} & \textbf{0.751} & \textbf{0.553} & \textbf{0.781} & \textbf{0.790} & $\checkmark$ & $\checkmark$\\
\bottomrule
\end{tabular}
\end{table}
\begin{table}[t]
\centering
\caption{Comparison with human performance. We estimate human performance by randomly splitting test-set annotators into two groups and treating their aggregated scores as ground truth and predictions, respectively. For fair comparison, VTON-IQA is evaluated against labels aggregated from half of the test-set annotators.}
\label{tab:compare_with_human}
\begin{tabular}{lrrrrr}
\toprule
Scorer 
& {$\rho_{\mathrm{SRCC}} \uparrow$}
& {$\rho_{\mathrm{PLCC}} \uparrow$}
& {$R^2 \uparrow$}
& {$A_{\rm macro} \uparrow$}
& {$A_{\rm micro} \uparrow$} \\
\midrule
Ours
& \num{0.699 +- 0.002}
& \num{0.700 +- 0.002}
& \num{0.489 +- 0.004}
& \num{0.771 +- 0.003}
& \num{0.783 +- 0.002} \\
Human
& {\bfseries \num{0.760 +- 0.004}}
& {\bfseries \num{0.762 +- 0.004}}
& {\bfseries \num{0.536 +- 0.008}}
& {\bfseries \num{0.782 +- 0.002}}
& {\bfseries \num{0.791 +- 0.002}} \\
\bottomrule
\end{tabular}
\end{table}
\begin{figure}[t]
  \centering
  \includegraphics[width=\linewidth]{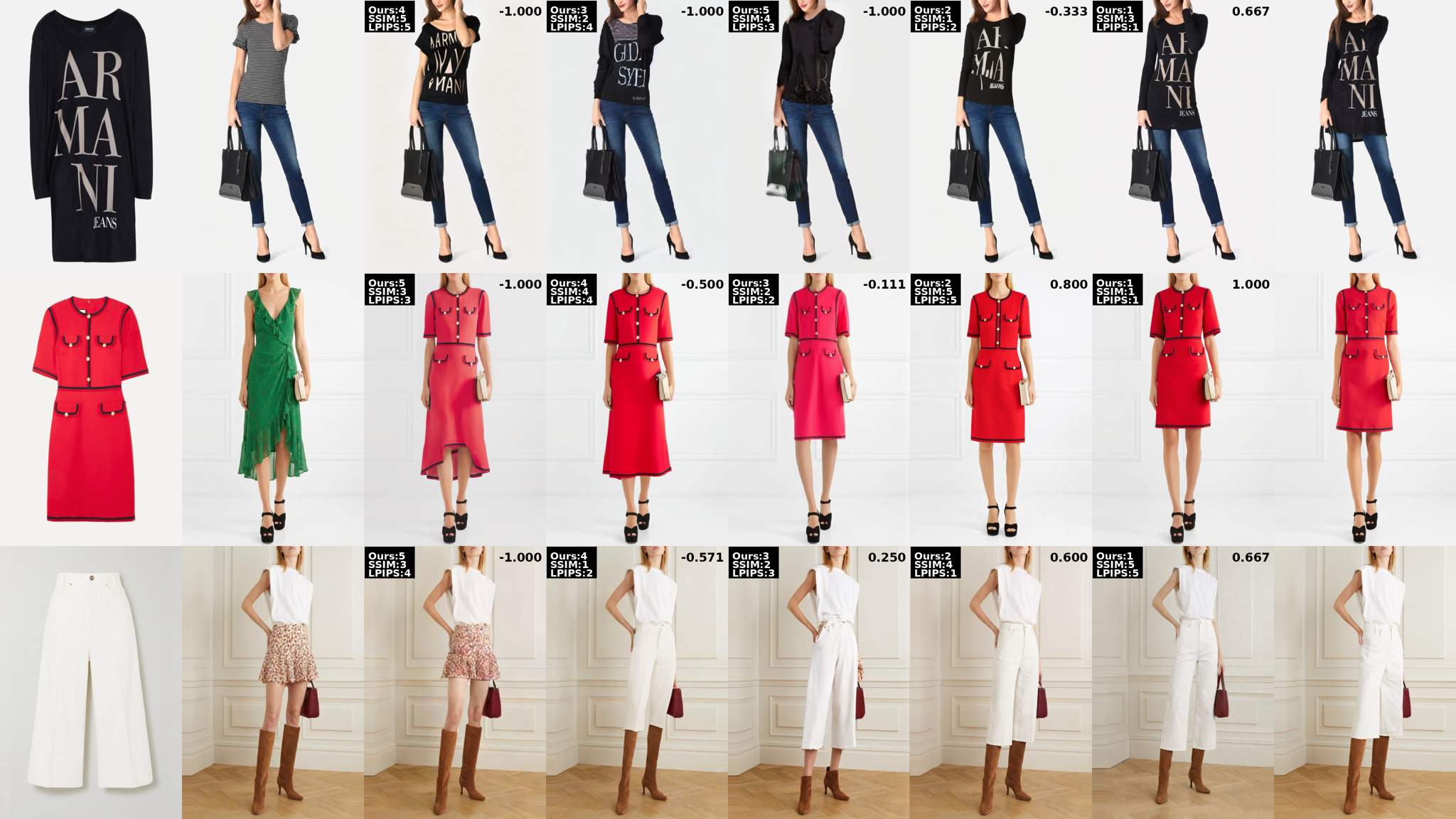}  \caption{Qualitative results. From left to right: garment image, target person image, generated try-on results (columns 3–7), and ground-truth image. The top-right value shows the human score, and the top-left black box indicates each metric’s ranking.}
  \label{fig:qualitative}
\end{figure}
\subsection{Quantitative Results}
\noindent\textbf{Comparison with Baselines.} Table~\ref{tab:compare_with_baseline} summarizes the evaluation results on the test set of VTON-QBench. For SSIM, LPIPS, and zero-shot DINOv3, $\rho_{\mathrm{PLCC}}$ and $R^{2}$ are not reported, as their score scales are not directly comparable to human subjective ratings, rendering these statistics less meaningful. Note that LPIPS is originally a ``lower-is-better'' metric; for consistency, we compute all correlation and regression metrics on its negated values so that all metrics follow a ``higher-is-better'' convention. Among reference-based metrics, SSIM exhibits weak alignment with human judgment, achieving low SRCC and pairwise accuracy. LPIPS performs better than SSIM but still shows a substantial gap compared to ours. The comparison between zero-shot DINOv3 and VTON-IQA without ICA highlights the importance of task-specific training on VTON-QBench, while further performance gains obtained by incorporating garment–person interaction modeling demonstrate the effectiveness of the proposed ICA module. The full VTON-IQA achieves the highest scores across all reported metrics, with $\rho_{\mathrm{SRCC}}=0.750$, $\rho_{\mathrm{PLCC}}=0.751$, and the best macro and micro pairwise accuracy.

\noindent\textbf{Comparison with Human.}
To estimate human performance, we split the annotators in the test set into two disjoint groups and compute independent quality scores from each group. One group is treated as ground-truth labels, while the other serves as predictions. This random partitioning is repeated 10 times, and we report the mean and standard deviation of the metrics across runs. For fair comparison with human performance, VTON-IQA is evaluated against the same ground-truth group in each run, while being trained on quality scores aggregated from all available annotators in the training set. Table~\ref{tab:compare_with_human} presents the comparison between human performance and our model. For correlation-based metrics ($\rho_{\mathrm{SRCC}}$ and $\rho_{\mathrm{PLCC}}$) and the coefficient of determination ($R^{2}$), a noticeable gap remains between our model and human performance, indicating room for further improvement in capturing fine-grained perceptual alignment. In contrast, our method achieves performance close to the human level in terms of macro accuracy ($A_{\rm macro}$) and micro accuracy ($A_{\rm micro}$). This suggests that, for pairwise quality comparisons of individual try-on images, the proposed model demonstrates human-comparable decision consistency.
\begin{table}[t]
\centering
\small
\caption{Evaluation results of VTON methods on Dress Code~\cite{dresscode}. 
Since several GAN-based models support only upper-body garments, their evaluation on Dress Code is restricted to upper-body categories and reported in \textcolor{gray!60}{gray} for reference. \textbf{Bold} and \underline{underlined} values indicate the best and second-best results for each metric, respectively.}
\label{tab:dresscode_paired_unpaired}
\begin{tabular}{
l
c c c c c
c c c
}
\toprule
& \multicolumn{5}{c}{\emph{paired}} & \multicolumn{3}{c}{\emph{unpaired}} \\
\cmidrule(lr){2-6} \cmidrule(lr){7-9}
Method
& {Ours $\uparrow$}
& {FID $\downarrow$}
& {KID $\downarrow$}
& {SSIM $\uparrow$}
& {LPIPS $\downarrow$}
& {Ours $\uparrow$}
& {FID $\downarrow$}
& {KID $\downarrow$} \\
\midrule
\multicolumn{9}{c}{\emph{Proprietary image edit model}} \\
\midrule
Nano Banana Pro~\cite{nanobanana}  & {\bfseries 0.305} & {\bfseries 1.837} & {\underline{0.471}} & {\underline{0.917}} & {\bfseries 0.045} & {\bfseries 0.295} & 5.179 & 1.048 \\
GPT-Image-1.5~\cite{gptimage}  & {\underline{0.237}} & 4.270 & 0.729 & 0.824 & 0.170 & {\underline{0.205}} & 5.621 & 0.845 \\
\midrule 
\multicolumn{9}{c}{\emph{DiT-based diffusion}} \\
\midrule
Qwen-Image-Edit~\cite{qwenedit} & -0.094 & 6.827 & 4.541 & {\bfseries 0.928} & 0.066 & -0.081 & 7.495 & 3.477 \\
CatVTON-FLUX~\cite{catvton}  & 0.014 & {\underline{3.093}} & {\bfseries 0.291} & 0.890 & 0.122 & -0.132 & {\underline{4.914}} & {\underline{0.767}} \\
FitDit~\cite{fitdit}  & 0.219 & 3.374 & 0.623 & 0.899 & 0.083 & 0.153 & {\bfseries 4.771} & {\bfseries 0.642} \\
Any2AnyTryon~\cite{any2any}  & -0.045 &  4.312 & 1.347 & 0.895 & 0.133 & -0.236 & 6.359 & 1.800 \\
\midrule 
\multicolumn{9}{c}{\emph{U-Net-based diffusion}} \\
\midrule
LADI-VTON~\cite{ladi}  & -0.562 & 6.608 & 2.907 & 0.893 & 0.150 & -0.694 & 9.131 & 4.414 \\
CAT-DM~\cite{catdm}  & -0.412 & 5.643 & 2.229 & 0.881 & 0.143 & -0.741 & 8.189 & 3.408 \\
OOTDiffusion~\cite{oot}  & -0.230 & 4.571 & 1.044 & 0.888 & 0.080 & -0.511 & 8.593 & 2.981 \\
IDM-VTON~\cite{idm} & 0.141  & 3.223 & 0.797 & 0.911 & {\underline{0.058}} & -0.349 & 4.931 & 1.095\\
CatVTON~\cite{catvton} & -0.081 & 5.306 & 1.841 & 0.882 & 0.110 & -0.232 & 6.927 & 2.015 \\
\midrule 
\multicolumn{9}{c}{\emph{GAN-based}} \\
\midrule
VITON-HD~\cite{vitonhd} & \textcolor{gray!60}{-0.933} & \textcolor{gray!60}{94.605} & \textcolor{gray!60}{90.826} & \textcolor{gray!60}{0.854} & \textcolor{gray!60}{0.224} & \textcolor{gray!60}{-0.933} & \textcolor{gray!60}{96.261} & \textcolor{gray!60}{94.204} \\
HR-VITON~\cite{hrviton} & \textcolor{gray!60}{-0.835} & \textcolor{gray!60}{20.873} & \textcolor{gray!60}{8.767} & \textcolor{gray!60}{0.917} & \textcolor{gray!60}{0.109} & \textcolor{gray!60}{-0.841} & \textcolor{gray!60}{22.626} & \textcolor{gray!60}{9.337} \\
SD-VITON~\cite{sdviton} & \textcolor{gray!60}{-0.868} & \textcolor{gray!60}{17.494} & \textcolor{gray!60}{5.736} & \textcolor{gray!60}{0.911} & \textcolor{gray!60}{0.107} & \textcolor{gray!60}{-0.874} & \textcolor{gray!60}{19.575} & \textcolor{gray!60}{6.798} \\
\bottomrule
\end{tabular}
\end{table}
\begin{table}[t]
\centering
\small
\caption{Evaluation results of VTON methods on VITON-HD~\cite{vitonhd}. \textbf{Bold} and \underline{underlined} values indicate the best and second-best results for each metric, respectively.}
\label{tab:vitonhd_paired_unpaired}
\begin{tabular}{
l
c c c c c
c c c
}
\toprule
& \multicolumn{5}{c}{\emph{paired}} & \multicolumn{3}{c}{\emph{unpaired}} \\
\cmidrule(lr){2-6} \cmidrule(lr){7-9}
Method
& {Ours $\uparrow$}
& {FID $\downarrow$}
& {KID $\downarrow$}
& {SSIM $\uparrow$}
& {LPIPS $\downarrow$}
& {Ours $\uparrow$}
& {FID $\downarrow$}
& {KID $\downarrow$} \\
\midrule
\multicolumn{9}{c}{\emph{Proprietary image edit model}} \\
\midrule
Nano Banana Pro~\cite{nanobanana}  & {\bfseries 0.303} & {\bfseries 3.318} & 0.707 & {\bfseries 0.904} & {\bfseries 0.052} & {\bfseries 0.315} & 10.309 & 2.454 \\
GPT-Image-1.5~\cite{gptimage}  & {\underline{0.255}} & 9.613 & 3.500 & 0.768 & 0.220 & {\underline{0.234}}
 & 12.801 & 4.631 \\
\midrule 
\multicolumn{9}{c}{\emph{DiT-based diffusion}} \\
\midrule
Qwen-Image-Edit~\cite{qwenedit} & 0.137 & 6.195 & 2.849 & 0.882 & {\underline{0.071}} & 0.087 & 10.706 & 3.030\\
CatVTON-FLUX~\cite{catvton}  & 0.207 & {\underline{5.470}} & {\underline{0.413}} & 0.872 & 0.132 & -0.025 & {\underline{9.084}} & {\underline{0.878}}\\
FitDit~\cite{fitdit}  & 0.230 & 8.048 & 1.232 & 0.843 & 0.144 & 0.189 & 9.893 & 1.355 \\
Any2AnyTryon~\cite{any2any}  & 0.074 & 9.043 & 3.271 & 0.860 & 0.169 & -0.082 & 12.018 & 3.892\\
\midrule 
\multicolumn{9}{c}{\emph{U-Net-based diffusion}} \\
\midrule
LADI-VTON~\cite{ladi}  & -0.756 & 15.404 & 7.416 & 0.847 & 0.202 & -0.864 & 21.515 & 12.333 \\
CAT-DM~\cite{catdm}  & -0.419 & 9.154 & 2.340 & 0.842 & 0.171 & -0.511 & 11.774 & 3.141 \\
OOTDiffusion~\cite{oot}  & 0.018 & 5.915 & {\bfseries 0.290} & 0.845 & 0.114 & -0.142 & {\bfseries 9.064} & {\bfseries 0.543} \\
IDM-VTON~\cite{idm} & 0.151 & 6.007 & 0.775 & 0.865 & 0.101 & 0.039 & 9.093 & 1.119 \\
CatVTON~\cite{catvton} & -0.163 & 9.082 & 2.710 & 0.833 & 0.196 & -0.266 & 11.199 & 2.824 \\
\midrule 
\multicolumn{9}{c}{\emph{GAN-based}} \\
\midrule
VITON-HD~\cite{vitonhd} & -0.512 & 9.450 & 1.956 & 0.877 & 0.130 & -0.559 & 11.535 & 2.437 \\
HR-VITON~\cite{hrviton} & -0.420 & 9.379 & 2.578 & 0.883 & 0.118 & -0.504 & 11.809 & 2.921 \\
SD-VITON~\cite{sdviton} & -0.368 & 7.746 & 1.567 & {\underline{0.886}} & 0.112 & -0.479 & 10.412 & 1.912 \\
\bottomrule
\end{tabular}
\end{table}
\subsection{Qualitative Results}
\label{sec:quality_assessment}
Fig.~\ref{fig:qualitative} presents representative qualitative comparisons. From left to right, the columns show the garment image, the target person image, the generated try-on results (columns 3–7), and the ground-truth image. The try-on results are arranged in ascending order of human subjective scores. For each try-on image, the value in the top-right corner denotes the human score, while the black box in the top-left corner indicates the rank of the five try-on images assigned by each metric. In the first row, only the top human-rated sample faithfully preserves both garment length and printed text details. Both our method and LPIPS correctly rank this sample as the top result, demonstrating sensitivity to fine-grained garment consistency. The second and third rows highlight cases involving structural changes. The second row contains pose variations, while the third row exhibits differences in zoom level, which often arise in mask-free try-on models without strict pose or scale constraints. Since SSIM and LPIPS rely on pixel- or feature-level alignment with the reference image, they tend to over-penalize such global transformations and produce rankings that deviate from human judgments. In contrast, VTON-IQA remains consistent with human perception, demonstrating robustness to pose and scale variations while still capturing try-on-specific quality differences.
\subsection{Quality Assessment Results for VTON Models}
We conduct a comprehensive evaluation of 14 representative VTON models under both \emph{paired} settings (where ground-truth try-on images are available) and \emph{unpaired} settings (where no reference images are provided). In the paired setting, reference-based metrics such as SSIM and LPIPS are applicable, whereas unpaired evaluation typically relies on distribution-level metrics including FID and KID. As a reference-free method, VTON-IQA can be applied consistently in both settings. All experiments are performed on the original test sets of Dress Code and VITON-HD, with results reported in Tables~\ref{tab:dresscode_paired_unpaired}--\ref{tab:vitonhd_paired_unpaired}. Across both datasets, proprietary image editing models achieve the highest VTON-IQA scores. Nano Banana Pro ranks first in all configurations, followed by GPT-Image-1.5, suggesting stronger alignment with human subjective ratings. In contrast, conventional full-reference metrics yield different rankings. Under SSIM and LPIPS, GPT-Image-1.5 performs substantially worse, particularly on VITON-HD, where it ranks below earlier GAN-based models. As discussed in Section~\ref{sec:quality_assessment}, these metrics penalize global structural variations such as pose and zoom changes. To further validate this hypothesis, we randomly sample 30 results per model and find that only one Nano Banana Pro sample exhibits noticeable pose or zoom variation, compared to 19 for GPT-Image-1.5, thereby accounting for the lower SSIM/LPIPS scores. More broadly, several diffusion models achieve competitive FID/KID scores yet rank lower than the proprietary models under VTON-IQA, indicating that distribution-level similarity does not necessarily reflect perceptual quality in virtual try-on.
\section{Conclusion}
We presented VTON-IQA, a reference-free, human-aligned image quality assessment framework for virtual try-on. To support reliable learning and evaluation, we constructed VTON-QBench, a large-scale human-annotated benchmark comprising 62,688 try-on images and 431,800 quality annotations from 13,838 qualified annotators. To ensure the reliability of these annotations, we developed a dedicated curation pipeline that quantitatively assesses annotator consistency and filters unreliable responses and annotators. This process provides reliable subjective supervision for training VTON-IQA to predict human-aligned quality scores from try-on, garment, and person images. By introducing an Interleaved Cross-Attention (ICA) module, VTON-IQA explicitly models interactions among these inputs, leading to stronger alignment with human subjective judgments. With VTON-IQA, we further conduct a comprehensive benchmark evaluation of 14 representative VTON models under both \emph{paired} and \emph{unpaired} settings, revealing that recent proprietary image-editing models achieve higher quality scores than OSS models. Meanwhile, conventional metrics such as SSIM, LPIPS, and FID/KID often fail to reflect these improvements, highlighting the gap between pixel- or distribution-level similarity and human-perceived try-on quality. Beyond our baseline benchmark, the public release of VTON-IQA can provide a fair, reproducible, and reference-free evaluation criterion for future VTON methods, serving as a scalable alternative to the ad hoc user studies often conducted in individual research projects. Overall, VTON-IQA and VTON-QBench support the standardization of quality assessment and sustained progress in the virtual try-on community.
\section*{Acknowledgements}
We thank Hideya Tanaka, Hiroshige Matsushita, and Motoaki Nakai
for their support in building and maintaining the data collection infrastructure.

\bibliographystyle{splncs04}
\bibliography{main}

@String(ECCV  = {Eur. Conf. Comput. Vis.})

@String(AAAI  = {AAAI})

@String(ECCV  = {ECCV})

@InProceedings{idm,
author="Choi, Yisol
and Kwak, Sangkyung
and Lee, Kyungmin
and Choi, Hyungwon
and Shin, Jinwoo",
title = {{Improving Diffusion Models for Authentic Virtual Try-on in the Wild}},
booktitle="Computer Vision -- ECCV 2024",
year="2025",
pages="206--235",
}

@article{vtbench,
  title={{VTBench: Comprehensive Benchmark Suite Towards Real-World Virtual Try-on Models}},
  author={Xiaobin Hu and Li Yujie and Donghao Luo and Pengcheng Xu and Jiangning Zhang and Junwei Zhu and Chengjie Wang and Yanwei Fu},
  journal={ArXiv},
  year={2025},
  volume={abs/2505.19571},
}

@inproceedings{vtonvllm,
 author = {Wan, Siqi and Chen, Jingwen and Cai, Qi and Pan, Yingwei and Yao, Ting and Mei, Tao},
 booktitle = {Advances in Neural Information Processing Systems},
 pages = {147641--147664},
 title = {{VTON-VLLM: Aligning Virtual Try-On Models with Human Preferences}},
 volume = {38},
 year = {2025}
}

@article{vtonqa,
  title={{VTONQA: A Multi-Dimensional Quality Assessment Dataset for Virtual Try-on}},
  author={Xinyi Wei and Sijing Wu and Zitong Xu and Yunhao Li and Huiyu Duan and Xiongkuo Min and Guangtao Zhai},
  journal={ArXiv},
  year={2026},
  volume={abs/2601.02945},
}

@inproceedings{qalign,
author = {Wu, Haoning and Zhang, Zicheng and Zhang, Weixia and Chen, Chaofeng and Liao, Liang and Li, Chunyi and Gao, Yixuan and Wang, Annan and Zhang, Erli and Sun, Wenxiu and Yan, Qiong and Min, Xiongkuo and Zhai, Guangtao and Lin, Weisi},
title = {{Q-ALIGN: Teaching LMMs for Visual Scoring via Discrete Text-Defined Levels}},
year = {2024},
booktitle = {Proceedings of the 41st International Conference on Machine Learning},
}

@inproceedings{clipiqa,
author = {Wang, Jianyi and Chan, Kelvin C.K. and Loy, Chen Change},
title = {{Exploring CLIP for Assessing the Look and Feel of Images}},
year = {2023},
pages = {2555--2563},
booktitle = {Proceedings of the AAAI Conference on Artificial Intelligence},
}

@InProceedings{any2any,
    author    = {Guo, Hailong and Zeng, Bohan and Song, Yiren and Zhang, Wentao and Liu, Jiaming and Zhang, Chuang},
    title     = {{Any2AnyTryon: Leveraging Adaptive Position Embeddings for Versatile Virtual Clothing Tasks}},
    booktitle = {Proceedings of the IEEE/CVF International Conference on Computer Vision},
    year      = {2025},
    pages     = {19085--19096}
}

@inproceedings{gpvton,
  title     = {{GP-VTON: Towards General Purpose Virtual Try-on via Collaborative Local-Flow Global-Parsing Learning}},
  author    = {Zhenyu, Xie and Zaiyu, Huang and Xin, Dong and Fuwei, Zhao and Haoye, Dong and Xijin, Zhang and Feida, Zhu and Xiaodan, Liang},
  booktitle = {Proceedings of the IEEE/CVF Conference on Computer Vision and Pattern Recognition},
  year      = {2023},
}

@inproceedings{sdviton,
  title={{Towards Squeezing-Averse Virtual Try-On via Sequential Deformation}},
  author={Shim, Sang-Heon and Chung, Jiwoo and Heo, Jae-Pil},
  booktitle={Proceedings of the AAAI Conference on Artificial Intelligence},
  pages={4856--4863},
  year={2024}
}

@inproceedings{oot,
  title={{OOTDiffusion: Outfitting Fusion based Latent Diffusion for Controllable Virtual Try-On}},
  author={Xu, Yuhao and Gu, Tao and Chen, Weifeng and Chen, Arlene},
  booktitle={Proceedings of the AAAI Conference on Artificial Intelligence},
  pages={8996--9004},
  year={2025}
}

@inproceedings{ladi,
author = {Morelli, Davide and Baldrati, Alberto and Cartella, Giuseppe and Cornia, Marcella and Bertini, Marco and Cucchiara, Rita},
title = {{LaDI-VTON: Latent Diffusion Textual-Inversion Enhanced Virtual Try-On}},
year = {2023},
booktitle = {Proceedings of the 31st ACM International Conference on Multimedia},
pages = {8580--8589},
}

@inproceedings{catvton,
title={{CatVTON: Concatenation Is All You Need for Virtual Try-On with Diffusion Models}},
author={Zheng Chong and Xiao Dong and Haoxiang Li and shiyue Zhang and Wenqing Zhang and Hanqing Zhao and xujie zhang and Dongmei Jiang and Xiaodan Liang},
booktitle={The Thirteenth International Conference on Learning Representations},
year={2025},
}

@article{fitdit,
  title={{FitDiT: Advancing the Authentic Garment Details for High-fidelity Virtual Try-on}},
  author={Boyuan Jiang and Xiaobin Hu and Donghao Luo and Qingdong He and Chengming Xu and Jinlong Peng and Jiangning Zhang and Chengjie Wang and Yunsheng Wu and Yanwei Fu},
  journal={ArXiv},
  year={2024},
  volume={abs/2411.10499},
}

@techreport{qwenedit,
  title={{Qwen-Image Technical Report}},
  author={Chenfei Wu and Jiahao Li and Jingren Zhou and Junyang Lin and Kaiyuan Gao and Kun Yan and Shengming Yin and Shuai Bai and Xiao Xu and Yilei Chen and Yuxiang Chen and Zecheng Tang and Zekai Zhang and Zhengyi Wang and An Yang and Bowen Yu and Chen Cheng and Dayiheng Liu and De-mei Li and Hang Zhang and Hao Meng and Hu Wei and Ji-Li Ni and Kai Chen and Kuang Cao and Liang Peng and Lin Qu and Min Wu and Peng Wang and Shuting Yu and Tingkun Wen and Wensen Feng and Xiao-Xue Xu and Yi Wang and Yichang Zhang and Yong-An Zhu and Yujian Wu and Yu-Jiao Cai and Ze-Yang Liu},
  journal={ArXiv},
  year={2025},
  institution={Qwen Team},
  volume={abs/2508.02324},
}

@misc{nanobanana,
  author = {Google},
  title = {{Nano Banana Pro}},
  year = {2025},
  note = {Accessed: 2026-02-11}
}

@InProceedings{catdm,
    author    = {Zeng, Jianhao and Song, Dan and Nie, Weizhi and Tian, Hongshuo and Wang, Tongtong and Liu, An-An},
    title     = {{CAT-DM: Controllable Accelerated Virtual Try-on with Diffusion Model}},
    booktitle = {Proceedings of the IEEE/CVF Conference on Computer Vision and Pattern Recognition},
    year      = {2024},
    pages     = {8372--8382}
}

@inproceedings{hrviton,
author = {Lee, Sangyun and Gu, Gyojung and Park, Sunghyun and Choi, Seunghwan and Choo, Jaegul},
title = {{High-Resolution Virtual Try-On with Misalignment and Occlusion-Handled Conditions
}},
year = {2022},
booktitle = {ECCV 2022: 17th European Conference, 2022, Proceedings},
pages = {204--219},
}

@misc{gptimage,
  author = {OpenAI},
  title = {{GPT-Image-1.5}},
  year = {2025},
  note = {Accessed: 2026-02-11}
}

@InProceedings{dresscode,
author="Morelli, Davide
and Fincato, Matteo
and Cornia, Marcella
and Landi, Federico
and Cesari, Fabio
and Cucchiara, Rita",
title = {{Dress Code: High-Resolution Multi-category Virtual Try-On}},
booktitle="Computer Vision -- ECCV 2022",
year="2022",
pages="345--362",
}

@inproceedings{vitonhd,
  title={{VITON-HD: High-Resolution Virtual Try-On via Misalignment-Aware Normalization}},
  author={Choi, Seunghwan and Park, Sunghyun and Lee, Minsoo and Choo, Jaegul},
  booktitle={Proceedings of the IEEE/CVF Conference on Computer Vision and Pattern Recognition},
  pages={14126--14135},
  year={2021}
}

@article{lhhuang2024iclora,
  title={{In-Context LoRA for Diffusion Transformers}},
  author={Lianghua Huang and Wei Wang and Zhigang Wu and Yupeng Shi and Huanzhang Dou and Chen Liang and Yutong Feng and Yu Liu and Jingren Zhou},
  journal={ArXiv},
  year={2024},
  volume={abs/2410.23775},
}

@misc{flux2024,
    author={{Black Forest Labs}},
    title={{FLUX}},
    year={2024},
    howpublished={\url{https://github.com/black-forest-labs/flux}},
}

@ARTICLE{ssim,
  author={Zhou Wang and Bovik, A.C. and Sheikh, H.R. and Simoncelli, E.P.},
  journal={IEEE Transactions on Image Processing}, 
  title={{Image quality assessment: from error visibility to structural similarity}}, 
  year={2004},
  volume={13},
  number={4},
  pages={600-612},
  }

@inproceedings{lpips,
  title={{The Unreasonable Effectiveness of Deep Features as a Perceptual Metric}},
  author={Zhang, Richard and Isola, Phillip and Efros, Alexei A and Shechtman, Eli and Wang, Oliver},
  booktitle={Proceedings of the IEEE/CVF Conference on Computer Vision and Pattern Recognition},
  year={2018}
}

@inproceedings{
kid,
title={Demystifying {MMD} {GAN}s},
author={Mikołaj Bińkowski and Dougal J. Sutherland and Michael Arbel and Arthur Gretton},
booktitle={The Sixth International Conference on Learning Representations},
year={2018},
}

@inproceedings{fid,
 author = {Heusel, Martin and Ramsauer, Hubert and Unterthiner, Thomas and Nessler, Bernhard and Hochreiter, Sepp},
 booktitle = {Advances in Neural Information Processing Systems},
 pages = {6629--6640},
 title = {{GANs Trained by a Two Time-Scale Update Rule Converge to a Local Nash Equilibrium}},
 volume = {30},
 year = {2017}
}

@article{dinov3,
  title={{DINOv3}}, 
  author={Oriane Siméoni and Huy V. Vo and Maximilian Seitzer and Federico Baldassarre and Maxime Oquab and Cijo Jose and Vasil Khalidov and Marc Szafraniec and Seungeun Yi and Michaël Ramamonjisoa and Francisco Massa and Daniel Haziza and Luca Wehrstedt and Jianyuan Wang and Timothée Darcet and Théo Moutakanni and Leonel Sentana and Claire Roberts and Andrea Vedaldi and Jamie Tolan and John Brandt and Camille Couprie and Julien Mairal and Hervé Jégou and Patrick Labatut and Piotr Bojanowski},
  journal={ArXiv},
  year={2025},
  volume={abs/2508.10104},
}

@article{bt,
 author = {Ralph Allan Bradley and Milton E. Terry},
 journal = {Biometrika},
 number = {3/4},
 pages = {324--345},
 publisher = {[Oxford University Press, Biometrika Trust]},
 title = {{Rank Analysis of Incomplete Block Designs: I. The Method of Paired Comparisons}},
 volume = {39},
 year = {1952}
}

@misc{mystylebox2026fashion,
  author       = {{Mystylebox}},
  title        = {{24 Types of Fashion Styles : Key Elements and Defining Looks}},
  howpublished = {https://www.mystylebox.com/pages/24-types-of-fashion-styles-explained},
  year         = {2026},
  note         = {Accessed: 2026-03-01}
}

@techreport{krippendorff,
  title={{Computing Krippendorff's Alpha-Reliability}},
  author={Krippendorff, Klaus},
  institution={University of Pennsylvania},
  year={2011}
}

@inproceedings{
adamw,
title={{Decoupled Weight Decay Regularization}},
author={Ilya Loshchilov and Frank Hutter},
booktitle={The Seventh International Conference on Learning Representations},
year={2019},
}

@InProceedings{viton,
author = {Han, Xintong and Wu, Zuxuan and Wu, Zhe and Yu, Ruichi and Davis, Larry S.},
title = {{VITON: An Image-Based Virtual Try-On Network}},
booktitle = {Proceedings of the IEEE Conference on Computer Vision and Pattern Recognition },
pages={7543--7552},
year = {2018}
}

@InProceedings{clothflow,
author = {Han, Xintong and Hu, Xiaojun and Huang, Weilin and Scott, Matthew R.},
title = {{ClothFlow: A Flow-Based Model for Clothed Person Generation}},
booktitle = {Proceedings of the IEEE/CVF International Conference on Computer Vision},
pages={10471--10480},
year = {2019}
}

@InProceedings{acgpn,
author = {Yang, Han and Zhang, Ruimao and Guo, Xiaobao and Liu, Wei and Zuo, Wangmeng and Luo, Ping},
title = {{Towards Photo-Realistic Virtual Try-On by Adaptively Generating-Preserving Image Content}},
booktitle = {Proceedings of the IEEE/CVF Conference on Computer Vision and Pattern Recognition},
year = {2020}
}

@inproceedings{pfafn,
  title={{Parser-Free Virtual Try-on via Distilling Appearance Flows}},
  author={Ge, Yuying and Song, Yibing and Zhang, Ruimao and Ge, Chongjian and Liu, Wei and Luo, Ping},
  booktitle={Proceedings of the IEEE/CVF Conference on Computer Vision and Pattern Recognition},
  pages={8481--8489},
  year={2021}
}

@inproceedings{gan,
 author = {Goodfellow, Ian J. and Pouget-Abadie, Jean and Mirza, Mehdi and Xu, Bing and Warde-Farley, David and Ozair, Sherjil and Courville, Aaron and Bengio, Yoshua},
 booktitle = {Advances in Neural Information Processing Systems},
 pages = {2672--2680},
 title = {{Generative Adversarial Nets}},
 volume = {27},
 year = {2014}
}

@inproceedings{ddpm,
 author = {Ho, Jonathan and Jain, Ajay and Abbeel, Pieter},
 booktitle = {Advances in Neural Information Processing Systems},
 pages = {6840--6851},
 title = {{Denoising Diffusion Probabilistic Models}},
 volume = {33},
 year = {2020}
}

@inproceedings{stablediffusion,
  title={{High-Resolution Image Synthesis with Latent Diffusion Models}},
  author={Rombach, Robin and Blattmann, Andreas and Lorenz, Dominik and Esser, Patrick and Ommer, Bj{\"o}rn},
  booktitle={Proceedings of the IEEE/CVF Conference on Computer Vision and Pattern Recognition},
  pages={10684--10695},
  year={2022}
}

@inproceedings{tryondiffusion,
  title={{TryOnDiffusion: A Tale of Two UNets}},
  author={Zhu, Luyang and Yang, Dawei and Zhu, Tyler and Reda, Fitsum and Chan, William and Saharia, Chitwan and Norouzi, Mohammad and Kemelmacher-Shlizerman, Ira},
  booktitle={Proceedings of the IEEE/CVF Conference on Computer Vision and Pattern Recognition},
  pages={4606--4615},
  year={2023}
}

@inproceedings{dit,
  title={{Scalable Diffusion Models with Transformers}},
  author={Peebles, William and Xie, Saining},
  booktitle={Proceedings of the IEEE/CVF International Conference on Computer Vision},
  pages={4195--4205},
  year={2023}
}

@inproceedings{
ku2024imagenhub,
title={{ImagenHub: Standardizing the evaluation of conditional image generation models}},
author={Max Ku and Tianle Li and Kai Zhang and Yujie Lu and Xingyu Fu and Wenwen Zhuang and Wenhu Chen},
booktitle={The Twelfth International Conference on Learning Representations},
year={2024},
}

@misc{ostris_ai_toolkit,
  author       = {Jaret Burkett},
  title        = {{AI Toolkit (ai-toolkit)}},
  howpublished = {\url{https://github.com/ostris/ai-toolkit/tree/main}},
  note         = {GitHub repository, accessed 2026-03-09},
  year         = {2026}
}

@inproceedings{openpose,
  author={Cao, Zhe and Simon, Tomas and Wei, Shih-En and Sheikh, Yaser},
  booktitle={Proceedings of the IEEE/CVF Conference on Computer Vision and Pattern Recognition}, 
  title={{Realtime Multi-Person 2D Pose Estimation Using Part Affinity Fields}}, 
  year={2017},
  pages={1302-1310}
  }

@inproceedings{cihp,
author = {Gong, Ke and Liang, Xiaodan and Li, Yicheng and Chen, Yimin and Yang, Ming and Lin, Liang},
title = {{Instance-Level Human Parsing via Part Grouping Network}},
year = {2018},
booktitle="Computer Vision -- ECCV 2018",
pages = {805--822},
}

@techreport{gpt4,
    author = {OpenAI},
    title = {{GPT-4 Technical Report}},
    institution = {OpenAI},
    year = {2023}
}

@misc{tensorflow,
title={ {TensorFlow: Large-Scale Machine Learning on Heterogeneous Systems}},
url={https://www.tensorflow.org/},
note={Software available from tensorflow.org},
author={
    Mart\'{i}n~Abadi and
    Ashish~Agarwal and
    Paul~Barham and
    Eugene~Brevdo and
    Zhifeng~Chen and
    Craig~Citro and
    Greg~S.~Corrado and
    Andy~Davis and
    Jeffrey~Dean and
    Matthieu~Devin and
    Sanjay~Ghemawat and
    Ian~Goodfellow and
    Andrew~Harp and
    Geoffrey~Irving and
    Michael~Isard and
    Yangqing Jia and
    Rafal~Jozefowicz and
    Lukasz~Kaiser and
    Manjunath~Kudlur and
    Josh~Levenberg and
    Dandelion~Man\'{e} and
    Rajat~Monga and
    Sherry~Moore and
    Derek~Murray and
    Chris~Olah and
    Mike~Schuster and
    Jonathon~Shlens and
    Benoit~Steiner and
    Ilya~Sutskever and
    Kunal~Talwar and
    Paul~Tucker and
    Vincent~Vanhoucke and
    Vijay~Vasudevan and
    Fernanda~Vi\'{e}gas and
    Oriol~Vinyals and
    Pete~Warden and
    Martin~Wattenberg and
    Martin~Wicke and
    Yuan~Yu and
    Xiaoqiang~Zheng},
  year={2015},
}

@inproceedings{pytorch,
 author = {Paszke, Adam and Gross, Sam and Massa, Francisco and Lerer, Adam and Bradbury, James and Chanan, Gregory and Killeen, Trevor and Lin, Zeming and Gimelshein, Natalia and Antiga, Luca and Desmaison, Alban and Kopf, Andreas and Yang, Edward and DeVito, Zachary and Raison, Martin and Tejani, Alykhan and Chilamkurthy, Sasank and Steiner, Benoit and Fang, Lu and Bai, Junjie and Chintala, Soumith},
 booktitle = {Advances in Neural Information Processing Systems},
 pages = {8026--8037},
 title = {{PyTorch: An Imperative Style, High-Performance Deep Learning Library}},
 volume = {32},
 year = {2019}
}

@misc{onnxruntime,
  title={{ONNX Runtime}},
  author={{ONNX Runtime developers}},
  year={2021},
  howpublished={\url{https://onnxruntime.ai/}},
}

\clearpage

\appendix
\section{Additional Details on VTON-QBench}
\subsection{Synthetic Pair Generation}
VTON-QBench is built on the test splits of VITON-HD~\cite{vitonhd} and Dress Code~\cite{dresscode}. However, the garment and person images contained in these test sets alone are insufficient in terms of diversity. 
To address this limitation, we expand the garment–person pair dataset across three categories: tops, bottoms, and dresses. Inspired by In-Context LoRA~\cite{lhhuang2024iclora}, we train a dedicated model $M_{\mathrm{ICL}}$ that generates a concatenated image 
$I_{\mathrm{G\parallel P}}\in\mathbb{R}^{3\times H\times 2W}$, which horizontally concatenates a garment image 
$I_{\mathrm{G}}\in\mathbb{R}^{3\times H\times W}$ and a person image 
$I_{\mathrm{P}}\in\mathbb{R}^{3\times H\times W}$ corresponding to an input prompt $T$. This model enables the generation of diverse garment–person pairs for data augmentation. The synthetic pair generation pipeline consists of four stages: (1) construction of the training dataset $\mathcal{D}_{\mathrm{ICL}}$, (2) training of the pair generation model $M_{\mathrm{ICL}}$, (3) synthetic pair generation, and (4) filtering of generated pairs. 
We describe each stage below.

\noindent\textbf{Construction of the Training Dataset $\mathcal{D}_{\mathrm{ICL}}$.}
The training dataset $\mathcal{D}_{\mathrm{ICL}}$ consists of paired samples of a concatenated image 
$I_{\mathrm{G\parallel P}}$ and its corresponding prompt $T$. 
For each of the three categories (upper-body, lower-body, dresses), we manually collected 40 high-quality garment–person pairs. 
To standardize the representation and facilitate efficient learning, we introduce a prompt template $\widehat{T}$ (Fig.~\ref{fig:base_prompt}) that explicitly describes a two-panel layout: a product image on the left and a person wearing the same garment on the right~\cite{lhhuang2024iclora}. 
For each pair, the prompt $T$ is constructed by filling the placeholders \texttt{<GARMENT\_DESCRIPTION>} and \texttt{<PERSON\_DESCRIPTION>} using GPT-4.1~\cite{gpt4}. The instruction used for this annotation process is shown in Fig.~\ref{fig:data_annotation_prompt}.

\noindent\textbf{Training of the Pair Generation Model $M_{\mathrm{ICL}}$.}
We train $M_{\mathrm{ICL}}$ using \textit{ai-toolkit}~\cite{ostris_ai_toolkit}, a widely used framework for diffusion model fine-tuning and LoRA training. 
This toolkit enables rapid prototyping by supporting various diffusion backbones with standardized training configurations. 
In our experiments, we fine-tune FLUX.1-dev~\cite{flux2024} following the default configuration provided in the toolkit repository as \texttt{train\_lora\_flux\_24gb.yaml}.

\noindent\textbf{Synthetic Pair Generation.}
To generate diverse garment–person pairs, careful prompt design is essential. 
We adopt a structured strategy to ensure both systematic style control and attribute diversity. 
First, we define target fashion styles based on established style categories (\eg, casual, street, formal, minimal, vintage) provided by Mystylebox~\cite{mystylebox2026fashion}. 
Next, for each garment category, we prepare a taxonomy of subcategories (Tab.~\ref{tab:garment_taxonomy}) and use GPT-4.1 to extract style-relevant subcategories via the prompt shown in Fig.~\ref{fig:category_association_prompt}. 
For each selected style–subcategory combination, we generate up to 50 diverse prompts using the template in Fig.~\ref{fig:prompt_generation_prompt}, enabling controlled variation in attributes, coordination, and styling. 
This process yields diverse yet systematically structured synthetic pairs.

\noindent\textbf{Filtering of Generated Pairs.}
Synthetic generation may introduce inconsistencies between the garment and person images in terms of category, color, shape, and texture. 
To ensure attribute consistency and visual validity, we introduce a three-stage filtering pipeline. 
First, we perform automatic filtering based on OpenPose~\cite{openpose}: 
(i) no person should be detected in the garment image, and 
(ii) exactly one front-facing person should be detected in the person image. 
Second, we use GPT to evaluate garment attribute consistency using the prompt shown in Fig.~\ref{fig:prompt_consistency_filtering_prompt}. 
Only pairs for which GPT outputs ``No critical mismatch detected.'' are retained. Finally, the remaining samples undergo manual inspection by 18 domain experts, consisting of nine practitioners who routinely handle fashion imagery and nine academic researchers in fashion-related fields. 
To ensure consistency, a unified evaluation guideline is established in advance and strictly followed during inspection. Through this process, approximately 60\% of the generated pairs are retained, ensuring the quality of the synthetic pairs.
\begin{figure}[t]
\centering
\begin{tcolorbox}
The two-panel image presents a product photo of a garment and a person modeling the garment; [LEFT] <GARMENT\_DESCRIPTION>; [RIGHT] <PERSON\_DESCRIPTION>.
\end{tcolorbox}
\caption{The prompt template for garment–person pair synthesis.}
\label{fig:base_prompt}
\end{figure}
\begin{figure}[t]
\centering
\begin{tcolorbox}
<PAIR\_IMAGE>\\
A combined image showing a product photo of a garment and a person modeling that garment is provided. Please create a prompt for generating this image following the format below:\\
The two-panel image presents a product photo of a garment and a person modeling the garment; [LEFT] <GARMENT\_DESCRIPTION>; [RIGHT] <PERSON\_DESCRIPTION>.
\end{tcolorbox}
\caption{The prompt for captioning garment–person pair images.}
\label{fig:data_annotation_prompt}
\end{figure}
\begin{figure}[t]
\centering
\begin{tcolorbox}
You are given a list of fashion style keywords: <FASHION\_STYLE>.\\
Below is a master list of basic fashion item categories: <CATEGORY\_MASTER>\\

From the list above, select and output only the fashion item categories that are strongly associated with the given fashion style keywords.
Consider typical garments, silhouettes, and combinations that define or frequently appear in that style.
Output the result as a bullet-point list of category names (in English, using only the given master list).\\

Requirements:\\
- Do NOT output any category that is only weakly or rarely related.\\
- Do NOT add any categories not present in the master list.\\
\end{tcolorbox}
\caption{The prompt for style category association.}
\label{fig:category_association_prompt}
\end{figure}
\begin{figure}[t]
\centering
\begin{tcolorbox}
You are given a garment category: <GARMENT\_CATEGORY>\\
The target fashion style is: <FASHION\_STYLE>\\

Generate 50 distinct prompts for an AI image generation model in bullet style.

Follow this format for the prompt: The two-panel image presents a product photo of a garment and a person modeling the garment; [LEFT] <GARMENT\_DESCRIPTION>; [RIGHT] <PERSON\_DESCRIPTION>.\\

Requirements:\\
- <GARMENT\_DESCRIPTION>: Describe the garment in detail as it appears in a product photo, inferring appropriate features and design elements that fit the specified fashion style (do not mention background, arrangement, or mannequins).\\
- <PERSON\_DESCRIPTION>: Describe a person wearing the exact same garment. Specify pose (front-facing or natural), styling (include tops, shoes, or accessories if relevant to the fashion style), and overall vibe matching the style.\\
- Each prompt must be unique, covering different settings, combinations, or style expressions within the given fashion style.\\
- Do not mention the garment category or any brand name in the prompt.\\
- Each bullet point should be a unique scenario.
\end{tcolorbox}
\caption{The prompt for generating prompts for garment–person pair synthesis.}
\label{fig:prompt_generation_prompt}
\end{figure}
\begin{figure}[t]
\centering
\begin{tcolorbox}
You are an expert in digital fashion evaluation.\\
You are provided with:\\
- An image of a standalone garment (product photo).\\
- An image of a person wearing what is supposed to be the exact same garment.\\

Your task is to determine whether the garment shown in both images is an exact match in terms of:\\
- Design features (shape, silhouette, collar, sleeves, hem, etc.)\\
- Color and pattern\\
- Material and texture\\
- Distinctive details (buttons, zippers, logos, prints, decorations, etc.)\\

Output your analysis as a bullet-point list.
Each item must be a single sentence, using 10 to 18 words, and state one critical mismatch or difference between the two garments.\\

If the garments are an exact match with no critical differences, output: "No critical mismatch detected."\\

Notes:\\
- Focus only on the garment itself.\\
- Do not mention the person, pose, body, or background.\\
\end{tcolorbox}
\caption{The prompt for garment–person pair consistency filtering.}
\label{fig:prompt_consistency_filtering_prompt}
\end{figure}
\begin{table*}[t]
\centering
\caption{Garment taxonomy used in this work.}
\label{tab:garment_taxonomy}
\begin{tabular}{p{3cm} p{8cm}}
\toprule
\textbf{Parent Category} & \textbf{Subcategories} \\
\midrule
Upper-body &
T-shirt, Long-sleeve, Blazer, Blouse, Cardigan,
Hoodie, Pullover, Shirt, Dress shirt, Flannel shirt,
Polo shirt, Sweater, Crewneck sweater, V-neck sweater,
Turtleneck, Sweatshirt, Tank top, Crop top, Vest,
Zip-up hoodie, Henley shirt. \\
\midrule
Lower-body &
Jeans, Straight-leg jeans, Skinny jeans, Bootcut jeans,
Wide-leg jeans, Cargo pants, Chinos, Slacks, Trousers,
Dress pants, Corduroy pants, Track pants, Sweatpants,
Joggers, Leggings, Biker shorts, Culottes, Capri pants,
Shorts, Cargo shorts, Denim shorts, Skirt, Mini skirt,
Midi skirt, Maxi skirt, Denim skirt, Pleated skirt,
Wrap skirt, Overalls, Palazzo pants, Flare pants, Skort. \\
\midrule
Dresses &
Dress, Mini dress, Midi dress, Maxi dress, Evening gown,
Cocktail dress, Slip dress, Bodycon dress, Sheath dress,
Wrap dress, Shirt dress, Tunic dress, Sundress,
Shift dress, A-line dress, Ball gown, Romper,
Jumpsuit, Playsuit, Halter dress, Strapless dress. \\
\bottomrule
\end{tabular}
\end{table*}

\subsection{Try-On Image Generation}
We generate virtual try-on images using 14 representative VTON models. 
For fair comparison, we use the official pretrained weights for each model and follow the inference configurations recommended by the original authors whenever available. Below, we describe the recomputation of clothing-agnostic representations for mask-based VTON models and the adaptation procedure for mask-free image-editing models.

\noindent\textbf{Recomputation of Clothing-Agnostic Representations.}
Different virtual try-on methods adopt distinct heuristics for generating clothing-agnostic representations. For fair comparison, we recompute the required human parsing and preprocessing for each person image and reconstruct the clothing-agnostic representation according to each method's specifications. In particular, VITON-HD originally generates clothing-agnostic representations using the Crowd Instance-level Human Parsing model~\cite{cihp}, whose official implementation is based on TensorFlow~\cite{tensorflow}. To integrate it into our PyTorch-based pipeline~\cite{pytorch}, we convert the parser to ONNX format~\cite{onnxruntime}.

\noindent\textbf{Adapting Mask-Free Image Editing Models to Virtual Try-On.} Recent advances in mask-free image editing models enable zero-shot virtual try-on without requiring clothing-agnostic representations. In this work, we incorporate virtual try-on images generated by these mask-free image editing models, including Qwen-Image-Edit~\cite{qwenedit}, Nano Banana Pro~\cite{nanobanana}, and GPT-Image-1.5~\cite{gptimage}, into VTON-QBench during its construction. To perform virtual try-on, we provide the person image as the primary input, the target garment image as a secondary input, and a textual instruction describing the desired editing operation. Fig.~\ref{fig:mask_free_vton_prompt} presents the prompt template used in our experiments. In the template, the placeholder \texttt{<CAPTION>} denotes a concise description (approximately 5–7 words) of the garment originally worn by the person. This description is automatically extracted using GPT to ensure consistent and structured prompts. The placeholder \texttt{<GARMENT\_CATEGORY>} specifies the category of the target garment, selected from \emph{top wear}, \emph{bottom wear}, or \emph{dress}.
\begin{figure}[t]
\centering
\begin{tcolorbox}
Replace the <CAPTION> worn by the person in image 1 with the <GARMENT\_CATEGORY> from image 2 while preserving the original person's pose, body shape, and identity, and ensuring the garment fits naturally on the body.
\end{tcolorbox}
\caption{The prompt used for performing virtual try-on with mask-free image editing models.}
\label{fig:mask_free_vton_prompt}
\end{figure}
\subsection{Details of the Crowdsourcing Procedure}
We recruited crowd workers via email invitations distributed to users of a fashion e-commerce platform. As compensation, participants were entered into a lottery for a gift voucher valued at approximately \$20.
\begin{figure}[t]
  \centering
  \includegraphics{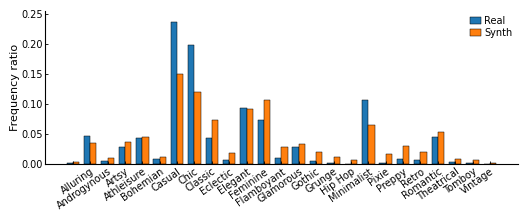}
   \caption{Distribution of fashion styles for real/synthetic pairs.}
   \label{fig:style}
\end{figure}
\begin{table}[t]
\centering
\small
\caption{Rows/columns denote train/test data, with S/R indicating synthetic/real pairs. Each cell reports $\rho_{\mathrm{SRCC}} / A$.}
\begin{tabular}{cccc}
\toprule
Train$\backslash$Test & S & R & R+S \\
\midrule
S   & $0.715 / 0.746$ & $0.595 / 0.755$ & $0.617 / 0.754$ \\
R   & $0.642 / 0.727 $ & $\mathbf{0.739} / 0.789 $ & $0.711 / 0.781 $  \\
R+S & $\mathbf{0.732} / \mathbf{0.751} $ & $0.738 / \mathbf{0.792}$ & $\mathbf{0.736} / \mathbf{0.787}$ \\
\bottomrule
\end{tabular}
\label{tab:synth_data_effectiveness}
\end{table}
\subsection{Effectiveness of Synthetic Pairs}
VTON-QBench contains 6,981 real and 6,172 synthetic pairs. To compare the attribute differences, we used GPT-5.4 to automatically annotate fashion style, age group, and body shape. As shown in Fig.~\ref{fig:style}, the synthetic pairs exhibit a longer-tailed style distribution than real ones, including styles that are extremely sparse in real data (\eg, Gothic, Grunge, and Pixie). Regarding the age group and body shape, 77\% are categorized as Adult (20--39) and 92\% as Slim in real data; for synthetic pairs, 88\% are categorized as Adult (20--39) and 95\% as Slim, revealing stronger biases than real data. 

To verify the effectiveness of synthetic pairs, we trained VTON-IQA with R (real only), S (synthetic only), and R+S (real+synthetic). For fairness, we randomly subsampled the real pairs to match the number of synthetic pairs. As shown in Tab.~\ref{tab:synth_data_effectiveness}, training with R+S improves performance on the broader R+S evaluation set without degrading performance on the real-only evaluation set R. This suggests that our three-stage filtering pipeline in Sec.~A.1, which removes roughly 60\% of the generated pairs, helps ensure the quality of the synthetic pairs.
\subsection{Effectiveness of Reference Image}
The perceived length of a garment in a VTON image can be ambiguous, as it depends not only on the image itself but also on factors such as the person’s height and the garment’s physical length. To mitigate this, our annotation interface includes a \emph{reference image} \(I_R\), showing the person actually wearing \(I_G\), along with \(I_P\), \(I_G\), and \(I_V\) (the top-right panel of Fig.~1). Thus, annotators judge relative consistency with \(I_R\), rather than absolute garment length, including whether the generated length appears visually plausible. For analysis, we used GPT-5.4 to identify VTON images with length-related issues and compared rating variance between samples with and without such issues. The standard deviations were 0.301 and 0.482 for 7,348 and 11,648 samples, respectively, showing even lower variance for those with length-related issues. This suggests that the \emph{reference image} supported consistent relative judgments. 

\section{Additional Experimental Results}
\subsection{Human–Model Calibration Analysis}
\begin{figure}[t]
  \centering
  \includegraphics[width=\linewidth]{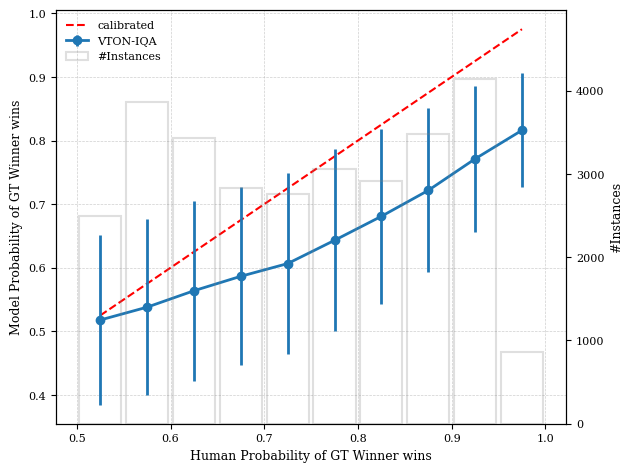}
  \caption{Human–Model calibration analysis. Human perceptual difficulty is defined based on pairwise comparisons between two virtual try-on results generated for the same garment–person pair. For each pair, we compute the probability that human annotators select the ground-truth (GT) image as the winner, following the procedure described in Sec.~4.2. These probabilities are grouped into bins with a width of 0.05. For each bin, we calculate the mean probability predicted by VTON-IQA that the GT image wins across all instances in the bin. The blue curve shows the mean model prediction, and error bars indicate the standard deviation within each bin. The gray bar plot (right y-axis) represents the number of instances per bin. The red dashed line denotes perfect calibration, where model predictions exactly match human preferences.}
  \label{fig:diff_analysis}
\end{figure}

Fig.~\ref{fig:diff_analysis} shows the calibration between human judgments and model predictions across different levels of perceptual difficulty. 
Human difficulty is defined based on pairwise comparisons between two virtual try-on results generated for the same garment--person pair, where we compute the probability that annotators select the ground-truth (GT) image as the winner, following the procedure described in Sec.~4.2. 
These probabilities are grouped into bins with a width of 0.05. 
For each bin, we report the mean probability predicted by VTON-IQA that the GT image wins. Error bars indicate the standard deviation within each bin, while the gray bars (right axis) represent the number of instances. 
The red dashed line denotes perfect calibration.

As shown in Fig.~\ref{fig:diff_analysis}, the model prediction exhibits a clear positive correlation with human perceptual difficulty. When even human annotators struggle to distinguish between the two images (\ie., the human preference probability is close to 0.5), the model prediction also remains near 0.5. 
As the task becomes easier for humans and the GT image is more consistently preferred, the predicted probability increases accordingly. Although the predictions are not perfectly calibrated, the consistent monotonic trend indicates that VTON-IQA behaves in line with human judgments. 
In particular, the model is uncertain when humans are uncertain and becomes more confident as the perceptual difference becomes clearer. 
This suggests that the model captures meaningful perceptual signals rather than relying on unintended shortcuts.
\subsection{Generalization to Unseen VTON Models}
To evaluate the generalization ability to unseen virtual try-on models, we conduct two complementary experiments: cross-group generalization and strong-model generalization. In the first experiment, we partition the VTON models into two groups and train VTON-IQA using only samples generated by models in one group. The trained model is then evaluated on the full test set, which includes samples generated by models from both groups. This design allows us to explicitly measure how well VTON-IQA generalizes to virtual try-on models that were not observed during training. In the second experiment, we simulate a more challenging scenario in which a new and stronger virtual try-on model emerges after the deployment of the quality assessment model. Specifically, we identify the top-performing models based on VTON-IQA scores obtained from full training, remove all corresponding instances from the training and validation split, and retrain the model from scratch. We then evaluate whether the retrained VTON-IQA can still correctly assess and rank these high-performing models when used for quality assessment. 
\begin{table}[t]
\centering
\begin{minipage}[t]{0.65\linewidth}
\centering
\caption{Ranking corr. \& coeff. of determination}
\label{tab:ranking_corr_and_r2}
\begin{tabular}{lccc|c}
\toprule
Cond.
& $\rho_{\mathrm{SRCC}} \uparrow$
& $\rho_{\mathrm{PLCC}} \uparrow$
& $R^2 \uparrow$ 
& $\bar{\Delta}_{\mathrm{rel}}$\\
\midrule
K
& \num{0.733+-0.016}
& \num{0.733+-0.012}
& \num{0.536+-0.017} 
& -\\
U
& \num{0.676+-0.018}
& \num{0.676+-0.019}
& \num{0.448+-0.025}
& \SI{10.6}{\percent}\\
K+U
& \num{0.704+-0.020}
& \num{0.704+-0.018}
& \num{0.492+-0.025}
& \SI{5.4}{\percent}\\
\bottomrule
\end{tabular}
\end{minipage}
\hfill
\begin{minipage}[t]{0.34\linewidth}
\centering
\caption{Macro Accuracy}
\label{tab:macro_acc}
\begin{tabular}{lc|c}
\toprule
Cond.
& $\bar{A}_{\rm macro} \uparrow$
& $\bar{\Delta}_{\mathrm{rel}}$\\
\midrule
KK
& \num{0.811+-0.025}
& -\\
UU
& \num{0.783+-0.006}
& \SI{3.5}{\percent}\\
KU
& \num{0.775+-0.013}
& \SI{4.4}{\percent}\\
\bottomrule
\end{tabular}
\end{minipage}
\end{table}
\begin{figure}[t]
  \centering
  \begin{subfigure}[t]{0.49\linewidth}
    \centering
    \includegraphics[width=\linewidth]{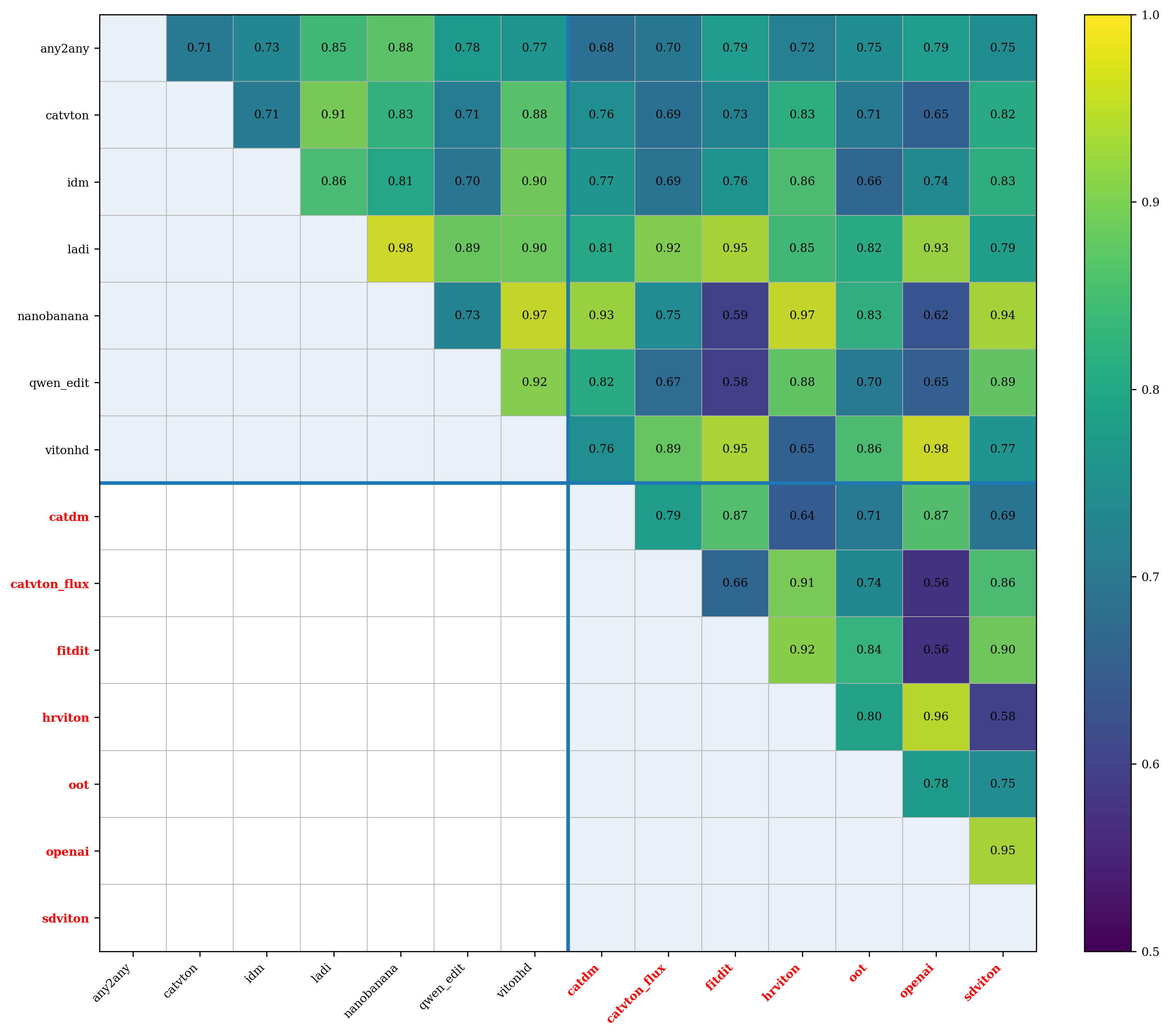}
    \caption{Group1.}
    \label{fig:generalization_heat_1}
  \end{subfigure}
  \hfill
  \begin{subfigure}[t]{0.49\linewidth}
    \centering
    \includegraphics[width=\linewidth]{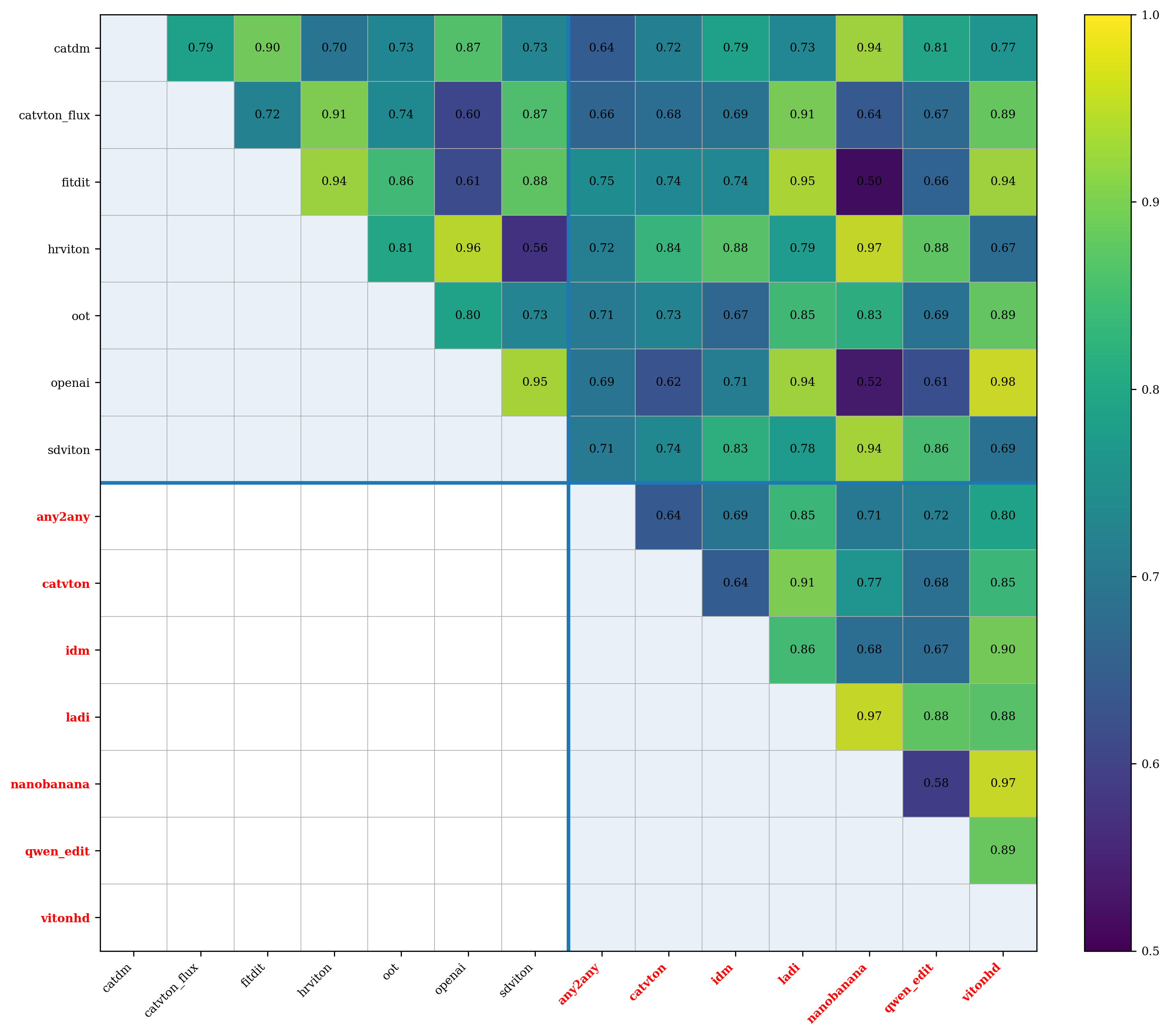}
    \caption{Group2.}
    \label{fig:generalization_heat_2}
  \end{subfigure}
  \caption{Pairwise accuracy matrix among virtual try-on models under the cross-group generalization setting. Each cell represents the micro accuracy obtained when comparing a pair of models on VTON-QBench. Rows and columns correspond to different virtual try-on models, and color intensity indicates the accuracy value. Models highlighted in red denote unknown models with respect to the training split. The upper-left block corresponds to comparisons between known models (KK), the lower-right block corresponds to comparisons between unknown models (UU), and the off-diagonal blocks (upper-right and lower-left) represent comparisons between known and unknown models (KU).}
  \label{fig:generalization_heat_gb}
\end{figure}
\begin{figure}[t]
  \centering
  \begin{subfigure}[t]{0.49\linewidth}
    \centering
    \includegraphics[width=\linewidth]{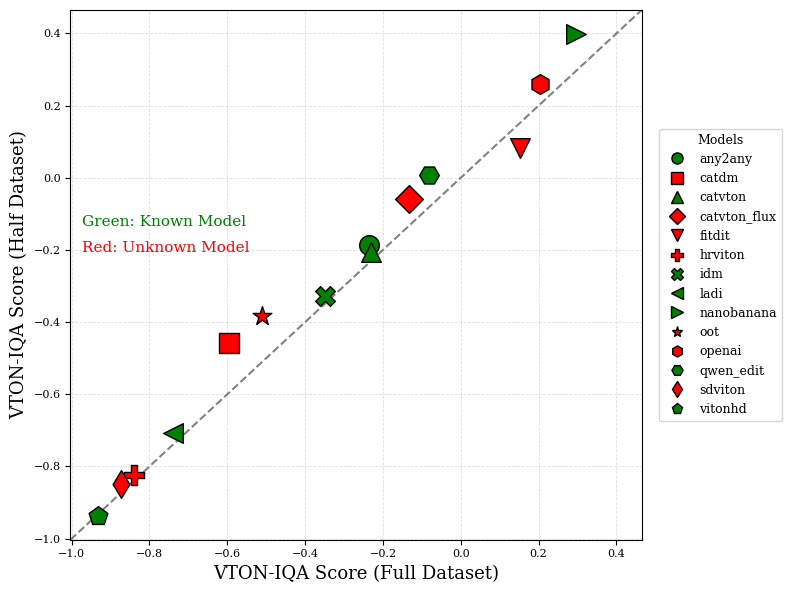}
    \caption{Group1.}
    \label{fig:generalization_1}
  \end{subfigure}
  \hfill
  \begin{subfigure}[t]{0.49\linewidth}
    \centering
    \includegraphics[width=\linewidth]{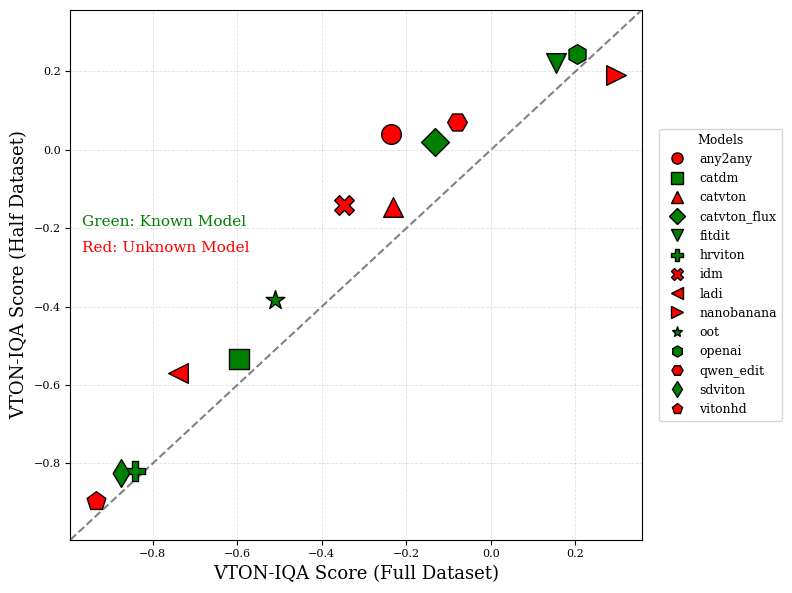}
    \caption{Group2.}
    \label{fig:generalization_2}
  \end{subfigure}
\caption{Comparison of VTON-IQA scores on the official unpaired test set of the Dress Code~\cite{dresscode} dataset under two training configurations. The x-axis shows the scores obtained when VTON-IQA is trained on the full VTON-QBench training set, while the y-axis shows the scores obtained when it is trained on a half of the dataset. Each marker corresponds to a VTON model; green and red denote known and unknown models, respectively. The dashed diagonal line indicates identical scores under the two training configurations.}
  \label{fig:generalization_scatter_1}
\end{figure}

\noindent\textbf{Cross-Group Generalization.} To evaluate the generalization capability of VTON-IQA to unseen virtual try-on models, we partition the 14 VTON models into two groups in a manner that maintains a balanced distribution of model performance across groups. Specifically, Group 1 (G1) comprises Any2AnyTryon~\cite{any2any}, CatVTON~\cite{catvton}, IDM-VTON~\cite{idm}, LADI-VTON~\cite{ladi}, Nano Banana Pro~\cite{nanobanana}, Qwen-Image-Edit~\cite{qwenedit}, and VITON-HD~\cite{vitonhd}. 
Group 2 (G2) includes CAT-DM~\cite{catdm}, CatVTON-FLUX~\cite{catvton}, FitDit~\cite{fitdit}, HR-VITON~\cite{hrviton}, OOTDiffusion~\cite{oot}, GPT-Image-1.5~\cite{gptimage}, and SD-VITON~\cite{sdviton}. We train VTON-IQA on the VTON-QBench training split restricted to samples generated by models in G1 and select the model based on validation performance on the validation split, also restricted to G1. The final evaluation is conducted on the full VTON-QBench test split, which includes samples generated by models from both G1 and G2. We then repeat the same procedure with the roles of G1 and G2 reversed.

Tab.~\ref{tab:ranking_corr_and_r2} reports the Spearman rank correlation coefficient (SRCC), Pearson linear correlation coefficient (PLCC), and coefficient of determination ($R^2$). The results are presented under three conditions: K (known models only), U (unknown models only), and K+U (all models). For Tab.~\ref{tab:ranking_corr_and_r2}, the relative performance drop $\bar{\Delta}_{\mathrm{rel}}$ is computed as the average decrease across the three metrics (SRCC, PLCC, and $R^2$), measured with respect to the known-only setting. Tab.~\ref{tab:macro_acc} reports the macro accuracy $\bar{A}_{\rm macro}$, derived from pairwise comparisons between virtual try-on images. For each pair of virtual try-on models, we first compute the micro accuracy based on the correctness of VTON-IQA’s pairwise preference predictions. The macro accuracy is then obtained by averaging these micro accuracies across all model pairs. The pairwise comparisons are categorized into three types: KK (comparisons between known models), UU (comparisons between unknown models), and KU (comparisons between known and unknown models). Within each category, the macro accuracy is computed by averaging the corresponding micro accuracies. In Tab.~\ref{tab:macro_acc}, $\bar{\Delta}_{\mathrm{rel}}$ denotes the relative decrease in macro accuracy with respect to the known-only setting. All reported values are presented as the mean and standard deviation over two runs, corresponding to training on G1 and training on G2, respectively.

For ranking-based metrics (Tab.~\ref{tab:ranking_corr_and_r2}), performance on unseen models (U) shows a relative decrease of 10.6\% compared to known models (K). However, when both known and unknown models are evaluated jointly (K+U), the relative drop is reduced to 5.4\%, indicating that the overall ranking consistency is largely preserved. Importantly, the SRCC remains above 0.67 even for unknown models, suggesting that relative ordering among models is still reasonably well maintained. In terms of macro accuracy (Tab.~\ref{tab:macro_acc}), the degradation is more limited. The relative drop is 3.5\% for comparisons between unknown models (UU) and 4.4\% for mixed comparisons (KU). This smaller decrease indicates that pairwise comparison accuracy remains relatively stable even in the presence of unknown models. Fig.~\ref{fig:generalization_heat_gb} provides a visualization of the pairwise accuracy matrix. In line with the quantitative results, comparable accuracy values are observed not only within the known-model block (KK) but also within the unknown-model block (UU) and across known–unknown comparisons (KU). 

Finally, to assess how the observed performance degradation on unknown models influences the final IQA outcomes, we apply VTON-IQA to the official unpaired test set of the Dress Code~\cite{dresscode} dataset. Fig.~\ref{fig:generalization_scatter_1} illustrates a comparison of VTON-IQA scores under the full- and half-dataset training settings. Most points are distributed close to the diagonal line, indicating that the relative ordering of models is largely preserved even when training is restricted to half of the model set. Notably, the points do not exhibit a consistent directional shift depending on whether a model is known or unknown. Although minor variations are observed for individual virtual try-on models, there is no clear evidence of systematic bias related to the known/unknown distinction. These results suggest that VTON-IQA can provide fair and consistent quality assessment even for virtual try-on models that were not included in the training data.

\noindent\textbf{Strong-Model Generalization.}
\begin{figure}[t] 
\centering 
\includegraphics[width=\linewidth]{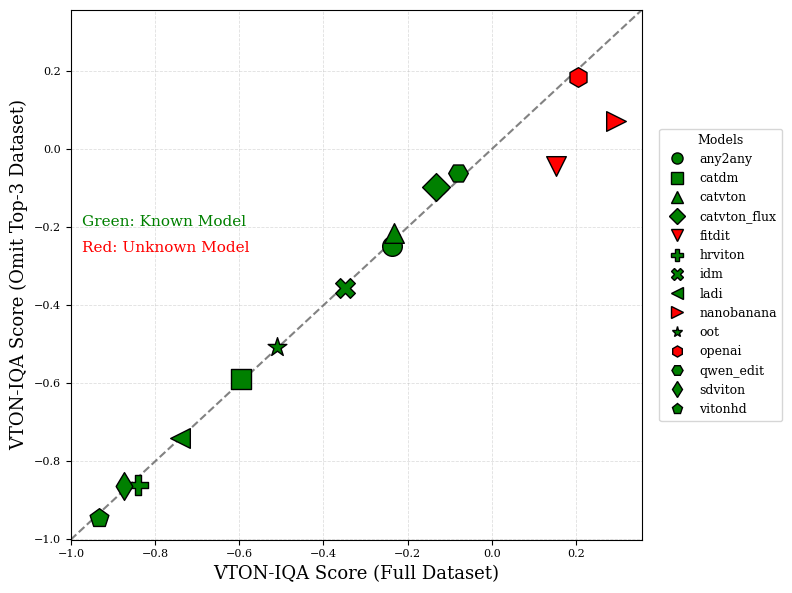} \caption{Comparison of VTON-IQA scores on the official unpaired test set of the Dress Code~\cite{dresscode} dataset under two training configurations. 
The x-axis shows the scores obtained when VTON-IQA is trained on the full VTON-QBench training set, while the y-axis shows the scores obtained when it is trained on the VTON-QBench dataset with all instances corresponding to the top-three models (Nano Banana Pro~\cite{nanobanana}, GPT-Image-1.5~\cite{gptimage}, and FitDit~\cite{fitdit}) removed. Each marker corresponds to a VTON model; green and red denote models included in and excluded from the retraining data, respectively. 
The dashed diagonal line indicates identical scores under the two training configurations.} 
\label{fig:generalization_3} 
\end{figure}
To evaluate whether VTON-IQA can still identify top-performing models without having observed them during training, we first train VTON-IQA on the full VTON-QBench training set and determine the top three virtual try-on models (Nano Banana Pro~\cite{nanobanana}, GPT-Image-1.5~\cite{gptimage}, and FitDit~\cite{fitdit}) according to its quality scores. We then remove all training instances corresponding to these three models from both the training and validation splits and retrain VTON-IQA from scratch. Finally, we examine whether the retrained model can still rank these models among the top performers when it is used as a quality assessment model.

Fig.~\ref{fig:generalization_3} compares VTON-IQA scores on the Dress Code~\cite{dresscode} unpaired test set when trained on the full dataset versus when trained with the top three models omitted. Although a slight change in ordering is observed between Nano Banana Pro and GPT-Image-1.5, the three excluded models remain ranked above the next-best model (Qwen-Image-Edit~\cite{qwenedit}). This suggests that VTON-IQA can assess the quality of future high-performing virtual try-on models beyond existing ones. For substantially stronger future models, a small set of annotated calibration samples can be added for fine-tuning or periodic retraining, without rebuilding the dataset.

\subsection{Exploring the Design Space of VTON-IQA Architectures}
\begin{table}[t]
\centering
\small
\caption{Comparison of different backbone types and training configurations.}
\label{tab:backbone_ft}
\begin{tabular}{llrrrrr}
\toprule
Backbone type & Training type & $\rho_{\mathrm{SRCC}} \uparrow$ & $\rho_{\mathrm{PLCC}} \uparrow$ & $R^2 \uparrow$ & {$A_{\rm macro} \uparrow$} & {$A_{\rm micro} \uparrow$} \\
\midrule
\multirow{3}{*}{DINOv3 ViT-B/16}
  & Full         & 0.222 & 0.242 & 0.055 & 0.592 & 0.579 \\
  & Last Half    & 0.726 & 0.724 & 0.521 & 0.770 & 0.784 \\
  & Last Quarter & 0.724 & 0.725 & 0.523 & 0.769 & 0.780 \\
\midrule
\multirow{3}{*}{DINOv3 ViT-L/16}
  & Full         & 0.226 & 0.254 & 0.063 & 0.609 & 0.590 \\
  & Last Half    & {\bfseries 0.750} & {\bfseries 0.751} & {\bfseries 0.553} &{\bfseries 0.781} & {\bfseries 0.790}\\
  & Last Quarter & 0.746 & 0.746 & {\bfseries 0.553} & 0.777 & 0.788 \\
\bottomrule
\end{tabular}
\end{table}
We conduct an ablation study to analyze the impact of backbone architecture and the number of fine-tuned layers in VTON-IQA. For the backbone, we compare DINOv3 ViT-B/16 and DINOv3 ViT-L/16. For the number of fine-tuned layers, we experiment with fine-tuning the full network, the last half of the layers, and the last quarter of the layers. As shown in Tab.~\ref{tab:backbone_ft}, fine-tuning the last 12 layers of DINOv3 ViT-L/16 achieves the best overall performance.
\begin{table}[t]
\centering
\small
\caption{Model-level rank correlation with human rankings on Dress Code and VTON-QBench.
Each cell reports $\mathring{\rho}_{\mathrm{SRCC}} / \bar{\rho}_{\mathrm{SRCC}}$.}
\begin{tabular}{ccccc}
\toprule
 & LPIPS & SSIM & FID & VTON-IQA \\
\midrule
Dress Code & \textemdash & \textemdash & \textemdash $/ \mathbf{0.991} $ & $\mathbf{0.772}  / 0.978 $ \\
QBench   & $0.483 / 0.367 $ & $ 0.239 /-0.218 $ & \textemdash $/ 0.930 $ & $\mathbf{0.736}  / \mathbf{0.987} $\\
\bottomrule
\end{tabular}
\label{tab:correlation}
\end{table}
\subsection{Further Correlation Analysis} To evaluate whether VTON-IQA reproduces human-based model rankings, we constructed an additional model-level ranking test set. We randomly sampled 30 person--garment pairs per category (upper, lower, and dresses) from the VTON-QBench and unpaired Dress Code test sets, with 14 VTON images per pair. Following Sec.~3.2, 22 reliable annotators participated, and each VTON image was evaluated by at least five annotators. Tab.~\ref{tab:correlation} reports two SRCC variants: \(\mathring{\rho}_{\mathrm{SRCC}}\), obtained by averaging the SRCC computed independently for each person--garment pair, and \(\bar{\rho}_{\mathrm{SRCC}}\), the SRCC computed on the model rankings obtained by averaging metric scores for each VTON model. Although FID achieves competitive \(\bar{\rho}_{\mathrm{SRCC}}\), it cannot assess per-pair quality. In contrast, VTON-IQA provides per-pair scores while maintaining strong alignment with human rankings.
\subsection{Category-wise Quality Assessment Results on Dress Code}
Tab.~\ref{tab:dresscode_paired_unpaired_upper_body}, 
Tab.~\ref{tab:dresscode_paired_unpaired_lower_body}, and 
Tab.~\ref{tab:dresscode_paired_unpaired_dresses} present the category-wise IQA results on the Dress Code dataset, covering upper-body garments, lower-body garments, and dresses, respectively. For each category, we report both paired and unpaired evaluation results, including our proposed VTON-IQA score and standard image quality metrics (FID~\cite{fid}, KID~\cite{kid}, SSIM~\cite{ssim}, and LPIPS~\cite{lpips}). 
\begin{table}[t]
\centering
\small
\caption{Evaluation results of VTON methods on the upper-body category of the Dress Code~\cite{dresscode} dataset.}
\label{tab:dresscode_paired_unpaired_upper_body}
\begin{tabular}{
l
c c c c c
c c c
}
\toprule
& \multicolumn{5}{c}{\emph{paired}} & \multicolumn{3}{c}{\emph{unpaired}} \\
\cmidrule(lr){2-6} \cmidrule(lr){7-9}
Method
& {Ours $\uparrow$}
& {FID $\downarrow$}
& {KID $\downarrow$}
& {SSIM $\uparrow$}
& {LPIPS $\downarrow$}
& {Ours $\uparrow$}
& {FID $\downarrow$}
& {KID $\downarrow$} \\
\midrule
\multicolumn{9}{c}{\emph{Proprietary image edit model}} \\
\midrule
Nano Banana Pro~\cite{nanobanana}  & {\bfseries 0.392} & {\bfseries 4.992} & {\bfseries 0.447} & 0.923 & {\bfseries 0.044} & {\bfseries 0.406}  & 12.127 & 1.341 \\
GPT-Image-1.5~\cite{gptimage}      & 0.334 & 10.105 & 1.012 & 0.837 & 0.163 & 0.323 & 13.337 & 1.403 \\
\midrule 
\multicolumn{9}{c}{\emph{DiT-based diffusion}} \\
\midrule
Qwen-Image-Edit~\cite{qwenedit}    & 0.055 & 10.101 & 4.355 & {\bfseries 0.940} & 0.052 & 0.150 & 14.782 & 4.525 \\
CatVTON-FLUX~\cite{catvton}        & 0.099 & 8.258 & {\underline{0.474}} & 0.914 & 0.103 & -0.005 & 11.863 & {\underline{0.855}} \\
FitDit~\cite{fitdit}               & {\underline{0.346}} & 8.275 & 0.699 & 0.924 & 0.062 & {\underline{0.328}} & {\underline{11.754}} & {\bfseries 0.750} \\
Any2AnyTryon~\cite{any2any}        & 0.131 & 9.001 & 1.646 & 0.892 & 0.142 & -0.026 & 13.007 & 2.112 \\
\midrule 
\multicolumn{9}{c}{\emph{U-Net-based diffusion}} \\
\midrule
LADI-VTON~\cite{ladi}              & -0.543 & 12.958 & 4.036 & 0.909 & 0.137 & -0.674 & 16.522 & 5.478 \\
CAT-DM~\cite{catdm}                & -0.082 & 10.416 & 2.389 & 0.909 & 0.119 & -0.335 & 14.181 & 3.275 \\
OOTDiffusion~\cite{oot}            & -0.064 & 8.684 & 0.660 & 0.908 & 0.066 & -0.232 & 13.504 & 1.834 \\
IDM-VTON~\cite{idm}                & 0.137 & {\underline{7.505}} & 1.055 & {\underline{0.928}} & {\underline{0.047}} & -0.082 & {\bfseries 11.512} & 1.291 \\
CatVTON~\cite{catvton}             & 0.057 & 9.938 & 1.764 & 0.905 & 0.084 & -0.021 & 12.914 & 1.800 \\
\midrule 
\multicolumn{9}{c}{\emph{GAN-based}} \\
\midrule
VITON-HD~\cite{vitonhd}            & -0.933 & 94.605 & 90.826 & 0.854 & 0.224 & -0.933 & 96.261 & 94.204 \\
HR-VITON~\cite{hrviton}            & -0.835 & 20.873 & 8.767 & 0.917 & 0.109 & -0.841 & 22.626 & 9.337 \\
SD-VITON~\cite{sdviton}            & -0.868 & 17.494 & 5.736 & 0.911 & 0.107 & -0.874 & 19.575 & 6.798 \\
\bottomrule
\end{tabular}
\end{table}

\begin{table}[t]
\centering
\small
\caption{Evaluation results of VTON methods on the lower-body category of the Dress Code~\cite{dresscode} dataset.}
\label{tab:dresscode_paired_unpaired_lower_body}
\begin{tabular}{
l
c c c c c
c c c
}
\toprule
& \multicolumn{5}{c}{\emph{paired}} & \multicolumn{3}{c}{\emph{unpaired}} \\
\cmidrule(lr){2-6} \cmidrule(lr){7-9}
Method
& {Ours $\uparrow$}
& {FID $\downarrow$}
& {KID $\downarrow$}
& {SSIM $\uparrow$}
& {LPIPS $\downarrow$}
& {Ours $\uparrow$}
& {FID $\downarrow$}
& {KID $\downarrow$} \\
\midrule
\multicolumn{9}{c}{\emph{Proprietary image edit model}} \\
\midrule
Nano Banana Pro~\cite{nanobanana}  & {\bfseries 0.279} & {\bfseries 4.554} & {\bfseries 0.341} & 0.904 & {\underline{0.050}} & {\bfseries{0.215}} & 13.201 & 1.680 \\
GPT-Image-1.5~\cite{gptimage}      & {\underline{0.214}} & 10.520 & 1.145 & 0.810 & 0.188 & {\underline{0.181}} & 13.913 & 1.225 \\
\midrule 
\multicolumn{9}{c}{\emph{DiT-based diffusion}} \\
\midrule
Qwen-Image-Edit~\cite{qwenedit}    & -0.315 & 19.233 & 11.014 & 0.913 & 0.088 & -0.222 & 15.682 & 5.082 \\
CatVTON-FLUX~\cite{catvton}        & 0.012 & 7.861 & {\underline{0.418}} & 0.896 & 0.121 & -0.166 & {\underline{12.338}} & {\bfseries 0.802} \\
FitDit~\cite{fitdit}               & 0.181 & 8.180 & 0.877 & 0.909 & 0.077 & 0.108 & 12.764 & {\underline{1.104}} \\
Any2AnyTryon~\cite{any2any}        & -0.110 & 9.439 & 1.571 & {\underline{0.914}} & 0.142 & -0.317 & 13.815 & 2.303 \\
\midrule 
\multicolumn{9}{c}{\emph{U-Net-based diffusion}} \\
\midrule
LADI-VTON~\cite{ladi}              & -0.420 & 9.540 & 1.987 & 0.907 & 0.140 & -0.697 & 13.966 & 3.480 \\
CAT-DM~\cite{catdm}                & -0.158 & 10.951 & 2.116 & 0.879 & 0.150 & -0.361 & 15.932 & 3.777 \\
OOTDiffusion~\cite{oot}            & -0.247 & 8.572 & 0.784 & 0.901 & 0.066 & -0.608 & 15.718 & 3.443 \\
IDM-VTON~\cite{idm}                & -0.010 & {\underline{7.256}} & 1.495 & {\bfseries 0.923} & {\bfseries 0.045} & -0.481 & {\bfseries 11.773} & 2.329 \\
CatVTON~\cite{catvton}             & -0.170 & 12.833 & 3.381 & 0.887 & 0.104 & -0.335 & 17.435 & 4.323 \\
\bottomrule
\end{tabular}
\end{table}

\begin{table}[t]
\centering
\small
\caption{Evaluation results of VTON methods on the dresses category of the Dress Code~\cite{dresscode} dataset.}
\label{tab:dresscode_paired_unpaired_dresses}
\begin{tabular}{
l
c c c c c
c c c
}
\toprule
& \multicolumn{5}{c}{\emph{paired}} & \multicolumn{3}{c}{\emph{unpaired}} \\
\cmidrule(lr){2-6} \cmidrule(lr){7-9}
Method
& {Ours $\uparrow$}
& {FID $\downarrow$}
& {KID $\downarrow$}
& {SSIM $\uparrow$}
& {LPIPS $\downarrow$}
& {Ours $\uparrow$}
& {FID $\downarrow$}
& {KID $\downarrow$} \\
\midrule
\multicolumn{9}{c}{\emph{Proprietary image edit model}} \\
\midrule
Nano Banana Pro~\cite{nanobanana}  & {\bfseries 0.243} & {\bfseries 3.147} & {\bfseries 0.412} & {\underline{0.924}} & {\bfseries 0.040} & {\bfseries 0.265} & {\bfseries 10.504} & 1.281 \\
GPT-Image-1.5~\cite{gptimage}      & {\underline{0.164}} & 8.180 & 0.805 & 0.825 & 0.158 & {\underline{0.108}} & 11.325 & 1.381 \\
\midrule 
\multicolumn{9}{c}{\emph{DiT-based diffusion}} \\
\midrule
Qwen-Image-Edit~\cite{qwenedit}    & -0.022 & 7.439 & 3.497 & {\bfseries 0.933} & {\underline{0.057}} & -0.171 & 13.062 & 3.720 \\
CatVTON-FLUX~\cite{catvton}        & -0.068 & 8.019 & {\underline{0.453}} & 0.861 & 0.142 & -0.225 & 11.524 & 1.499 \\
FitDit~\cite{fitdit}               & 0.130 & 7.886 & 0.926 & 0.864 & 0.111 & 0.023 & 10.767 & {\bfseries 1.211} \\
Any2AnyTryon~\cite{any2any}        & -0.157 & 8.644 & 1.443 & 0.879 & 0.141 & -0.366 & 13.190 & 2.788 \\
\midrule 
\multicolumn{9}{c}{\emph{U-Net-based diffusion}} \\
\midrule
LADI-VTON~\cite{ladi}              & -0.724 & 15.434 & 5.023 & 0.863 & 0.174 & -0.851 & 20.363 & 8.295 \\
CAT-DM~\cite{catdm}                & -0.404 & 11.896 & 2.888 & 0.855 & 0.159 & -0.680 & 16.585 & 5.417 \\
OOTDiffusion~\cite{oot}            & -0.379 & 12.306 & 2.516 & 0.856 & 0.108 & -0.693 & 19.963 & 6.146 \\
IDM-VTON~\cite{idm}                & -0.026 & {\underline{7.278}} & 0.619 & 0.882 & 0.083 & -0.480 & {\underline{11.303}} & {\underline{1.220}} \\
CatVTON~\cite{catvton}             & -0.130 & 10.394 & 2.169 & 0.887 & 0.104 & -0.339 & 13.091 & 2.810 \\
\bottomrule
\end{tabular}
\end{table}

\subsection{Qualitative Results}
\begin{figure}[t]
  \centering
  \includegraphics[width=\linewidth]{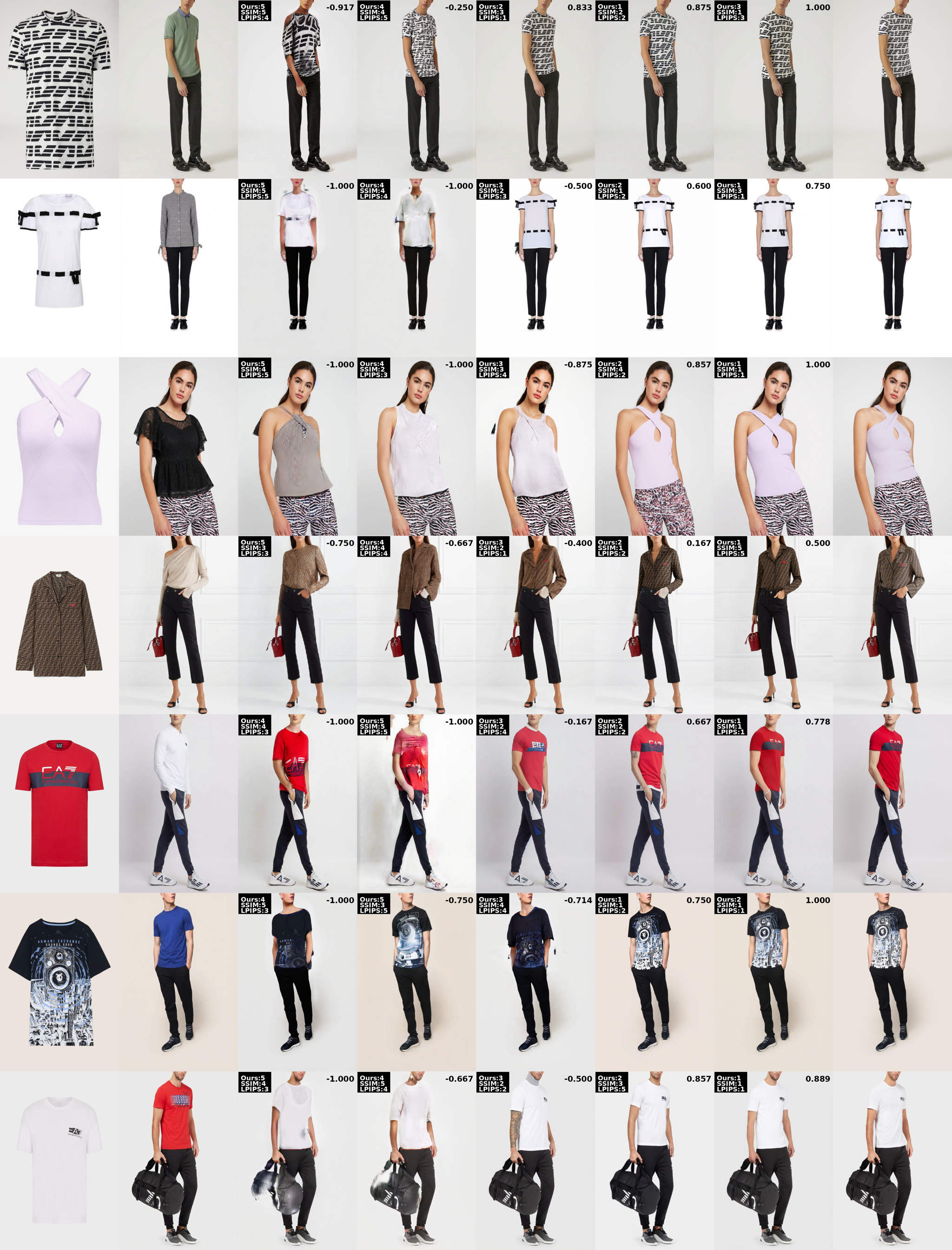}
  \caption{Qualitative results. From left to right: garment image, target person image, generated try-on results (columns 3–7), and ground-truth image. The top-right value shows the human score, and the top-left black box indicates each metric’s ranking.}
  \label{fig:qualitative_ub}
\end{figure}

\begin{figure}[t]
  \centering
  \includegraphics[width=\linewidth]{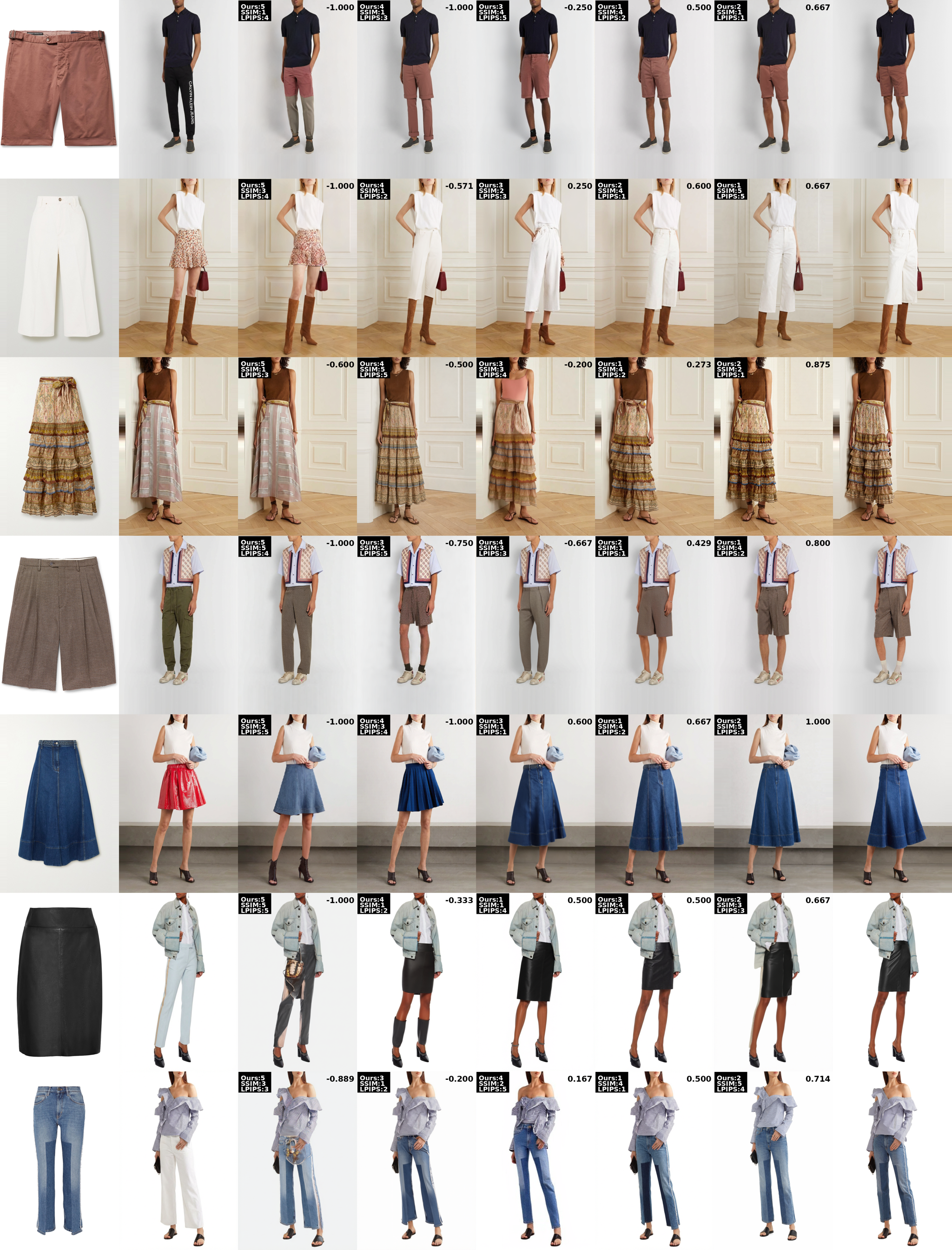}
  \caption{Qualitative results. From left to right: garment image, target person image, generated try-on results (columns 3–7), and ground-truth image. The top-right value shows the human score, and the top-left black box indicates each metric’s ranking.}
  \label{fig:qualitative_lb}
\end{figure}

\begin{figure}[t]
  \centering
  \includegraphics[width=\linewidth]{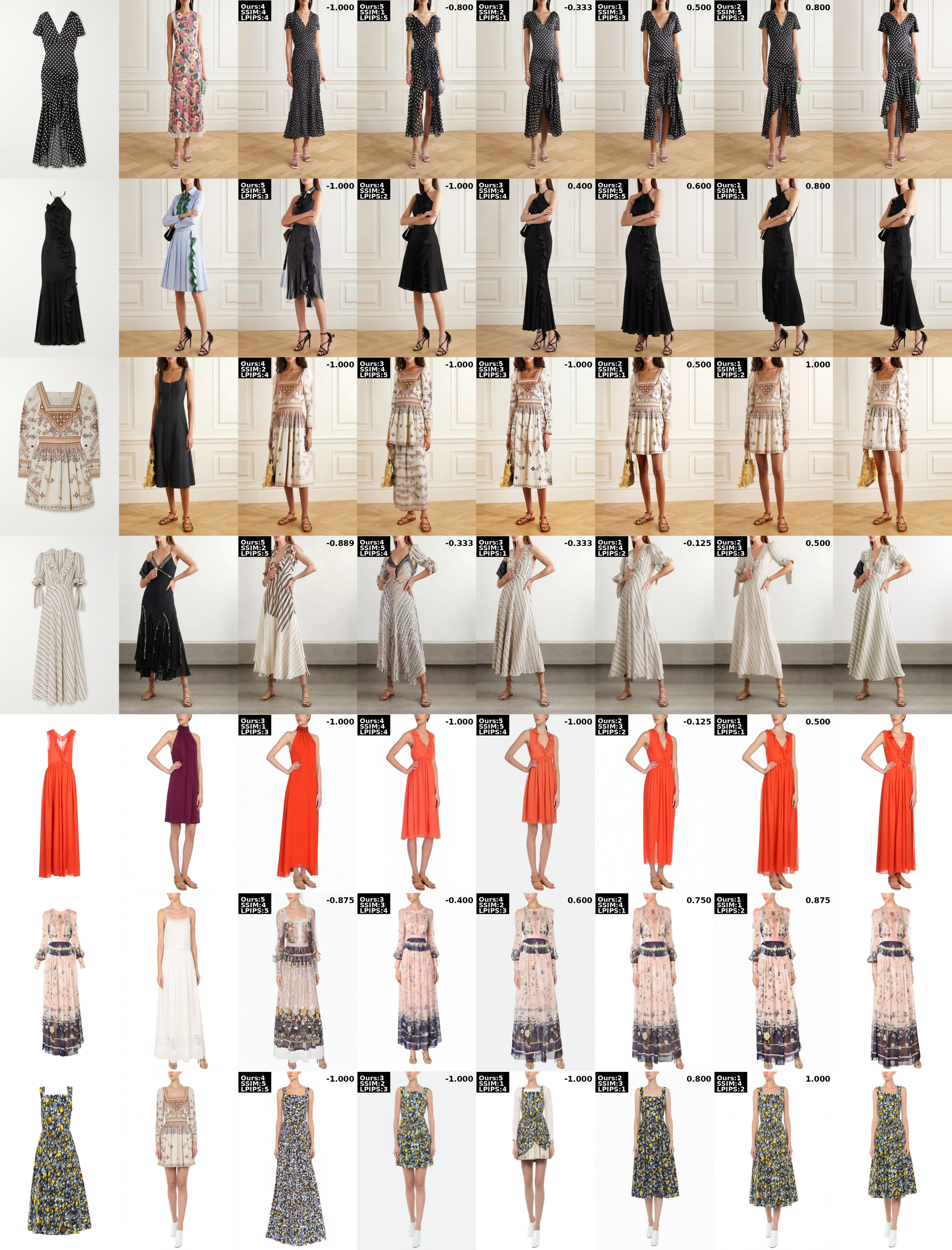}
  \caption{Qualitative results. From left to right: garment image, target person image, generated try-on results (columns 3–7), and ground-truth image. The top-right value shows the human score, and the top-left black box indicates each metric’s ranking.}
  \label{fig:qualitative_dr}
\end{figure}

We present further visualizations of VTON-IQA quality assessment results in Fig.~\ref{fig:qualitative_ub}, Fig.~\ref{fig:qualitative_lb}, and Fig.~\ref{fig:qualitative_dr}.
\section{Limitations \& Future Work}
VTON-IQA is capable of assessing the quality of individual virtual try-on images even in the absence of ground-truth references. This property makes it particularly suitable for real-world applications where such references are unavailable.

However, several limitations remain. First, our current study focuses on standard studio-based scenarios with controlled garment–person pairs. Extending the proposed framework to more diverse and realistic conditions, such as in-the-wild images with complex backgrounds and poses, person-to-person transfer, and instruction-guided styling, offers an exciting direction for future research. Second, our framework emphasizes scalar quality score prediction for quantitative evaluation. While effective for benchmarking, it does not yet provide detailed and interpretable feedback about specific visual discrepancies, such as sleeve length, silhouette, or fine-grained design elements. Incorporating attribute-level reasoning or language-based explanations into the evaluation process could enhance interpretability and practical usability, representing a promising avenue for future work. Third, our framework is currently designed for image-based virtual try-on and does not explicitly address video-based or 3D settings. Extending the evaluation paradigm to these emerging scenarios would require modeling temporal consistency and geometric plausibility. Adapting the framework to capture such factors could further advance human-aligned evaluation of next-generation virtual try-on systems. Despite these limitations, our method provides a principled and empirically validated foundation for human-aligned evaluation of virtual try-on systems and can serve as a basis for these future extensions.
\clearpage
\end{document}